\documentclass[twoside]{article}

\usepackage[accepted]{aistats2025}
\usepackage[utf8]{inputenc} 
\usepackage[T1]{fontenc}    
\usepackage[backref=page]{hyperref}       
\usepackage{url}   

\usepackage{booktabs}       
\usepackage{amsfonts}       
\usepackage{nicefrac}       
\usepackage{microtype}      
\usepackage{xcolor}         
\usepackage{mathtools}
\usepackage[symbol]{footmisc}
\usepackage{algpseudocode}
\usepackage{algorithm}
\usepackage{enumitem}
\usepackage{float}
\usepackage{amsmath,amsthm}
\usepackage{tikz-cd}
\usepackage{tikzit}

\usepackage{wrapfig}
\usepackage{authblk}
\usepackage{graphicx, tikz, graphics, epsf, caption, subcaption}
\usepackage{multirow}
\usetikzlibrary{arrows, positioning, backgrounds} 
\usetikzlibrary{bayesnet}
\newtheorem{definition}{Definition}[section]

\newtheorem{proposition}{Proposition}[section]

\def\EE{{\mathbb E}}    
\def\11{{\mathbf 1}}    

\def\cA{{\mathcal A}}     \def\cB{{\mathcal B}}   \def\cN{{\mathcal N}} \def\cT{{\mathcal T}}   \def\cI{{\mathcal I}} \def\cO{{\mathcal O}}  \def\cD{{\mathcal D}}              \def\cX{{\mathcal X}} \def\cY{{\mathcal Y}}  
  
\def\cI{{\mathcal I}}  
\def\causalparam{{\theta_{\mathrm{c}}}}  

            \def\bm{{\mathbf m}}       \def\bt{{\mathbf t}}    \def\bx{{\mathbf x}} \def\by{{\mathbf y}}  \def\bt{{\mathbf t}}
\def\bX{{\mathbf X}} 
\def\bA{{\mathbf A}}    \def\bE{{\mathbf E}}      \def\bK{{\mathbf K}}       \def\bR{{\mathbf R}}      \def\bX{{\mathbf X}}

\DeclareSymbolFont{boldoperators}{OT1}{cmr}{bx}{n}
\SetSymbolFont{boldoperators}{bold}{OT1}{cmr}{bx}{n}

\newcommand{\norm}[1]{\left\lvert #1 \right\rvert}
\newcommand{\Norm}[1]{\left\lVert #1 \right\rVert}

\newcommand{\circbrac}[1]{\left({#1}\right)}

\usepackage{parskip}

\usetikzlibrary{shapes, arrows.meta}

\tikzstyle{box}=[fill={rgb,255: red,128; green,128; blue,128}, draw=black, shape=rectangle, tikzit fill={rgb,255: red,128; green,128; blue,128}, tikzit shape=rectangle, minimum width=3em]
\tikzstyle{new style 0}=[fill=none, draw=black, shape=rectangle, minimum width=15em, minimum height=10em]
\tikzstyle{Cylinder}=[fill=white, draw=black, shape=cylinder, rotate=90, anchor=north, minimum width=3.5 em]

\tikzstyle{Arrow}=[->, fill=none]

\tikzset{rv/.style={circle, draw, thick, minimum size=6mm, inner sep=0.5mm}}
\tikzset{rve/.style={ellipse, draw, thick, minimum size=6mm, inner sep=0.3mm}}
\tikzset{fv/.style={rectangle, draw, thick, minimum size=6mm, inner sep=0.5mm}}
\tikzset{rvl/.style={circle, red, draw, thick, minimum size=6mm, inner sep=0.3mm}}
\tikzset{rv0/.style={circle, draw, thick, minimum size=6mm, inner sep=0.3mm}}
\tikzset{rv1/.style={circle, draw, minimum size=6mm, inner sep=0.3mm}}
\tikzset{deg/.style={->, very thick}}
\tikzset{degl/.style={->, very thick, color=red}}
\tikzset{begl/.style={<->, very thick, color=red, dashed}}
\tikzset{beg/.style={<->, very thick, color=red}}
\def\equationautorefname#1#2\null{%
  Eq.#1#2\null%
}
\usepackage[round]{natbib}

\definecolor{darkblue}{RGB}{0, 0, 128}
\hypersetup{
	colorlinks=true,
	linkcolor=darkblue,
	filecolor=darkblue,
	urlcolor=darkblue,
	citecolor=darkblue,
	pdfauthor={},
	pdftitle={}
}

\def\equationautorefname#1#2\null{%
  Eq.\!#1#2\null%
}

\begin{document}

%

%

\twocolumn[

\aistatstitle{Is Merging Worth It? Securely Evaluating the Information Gain for Causal Dataset Acquisition}

\aistatsauthor{ Jake Fawkes\footnotemark[1] \And Lucile Ter-Minassian\footnotemark[1] }
\aistatsaddress{ 
    University of Oxford 
    \And    
    University of Oxford 
}

\aistatsauthor{ Desi R. Ivanova \And Uri Shalit \And Chris Holmes}

\aistatsaddress{University of Oxford \And Technion - Israel \\ Institute of Technology \And  University of Oxford} ]

\runningauthor{Jake Fawkes*, Lucile Ter-Minassian*, Desi R. Ivanova, Uri Shalit , Chris Holmes}


\footnotetext[1]{  Correspondence to: \texttt{jake.fawkes@st-hughs.ox.ac.uk} and \texttt{lucile.ter-minassian@spc.ox.ac.uk}.}

\begin{abstract}
Merging datasets across institutions is a lengthy and costly procedure, especially when it involves private information. 
Data hosts may therefore want to prospectively gauge which datasets are most beneficial to merge with, without revealing sensitive information. For causal estimation this is particularly challenging as the value of a merge depends not only on reduction in epistemic uncertainty but also on improvement in overlap. 
To address this challenge, we introduce the first \emph{cryptographically secure} information-theoretic approach for quantifying the value of a merge in the context of heterogeneous treatment effect estimation. 
We do this by evaluating the \emph{Expected Information Gain} (EIG) using multi-party computation to ensure that no raw data is revealed. 
We further demonstrate that our approach can be combined with differential privacy (DP) to meet arbitrary privacy requirements whilst preserving more accurate computation compared to DP alone. 
To the best of our knowledge, this work presents the first privacy-preserving method for dataset acquisition tailored to causal estimation.
Code is publicly available: \url{https://github.com/LucileTerminassian/causal_prospective_merge}.

\end{abstract}

\section{INTRODUCTION}

As the demand for data-driven decision making grows, the question of how to optimally collect data for a given task becomes increasingly important.
Data \hspace{0em} fusion~\citep{castanedo2013review}, which integrates pre-existing data from various sources, is a popular method to increase sample size, reduce sampling variability, and enhance statistical power and robustness \citep{lenzerini2002data, doan2012principles}.
However, merging datasets is often a time-consuming and resource-intensive task. This is especially true in sensitive domains such as healthcare, where concerns surrounding privacy, downstream applications, and data security mean long ethical approval procedures are required before undergoing a merge~\citep{platt2015public, mello2013preparing, europeancommission2018dataprotection}. 
Consequently, practitioners crucially need methods to determine the value of a potential merge in advance, whilst also complying with privacy requirements \citep{abouelmehdi2018big}.

In this work, we focus on data fusion in the context of heterogeneous treatment effect estimation. As a concrete example of this problem, consider a hospital that wishes to assess the impact of a medical intervention on its patients. Upon finding that its own data is insufficient to get accurate estimates, the hospital plans to select one of $K$ possible candidate hospitals for a potential data merge. 

Given the costs involved, the hospital would like to identify \emph{in advance} which potential dataset would provide the most information upon merging, \emph{whilst} complying with privacy regulations. We propose a solution for such types of problems, allowing for scenarios where patient outcomes at the candidate hospitals to be unobserved or simply masked. 
\begin{figure*}[h]
\centering
    \vspace{10pt}
    \begin{tikzpicture}[scale = 0.4]
	\begin{pgfonlayer}{nodelayer}
		\node [style=Cylinder] (0) at (-10.75, 2.75) {\rotatebox{270}{$\mathcal{D}_0$}};
		\node [style=none] (3) at (-9.5, 0.25) {\begin{tabular}{c}
    1.  Define $f_{\theta},P(\theta)$ \\  and  fit to $\mathcal{D}_0$.\end{tabular}};
		\node [style=none] (5) at (-9.5, 6.25) {\textbf{Host site}};
		\node [style=Cylinder] (7) at (6.5, 2.75) {\rotatebox{270}{$\mathcal{D}_1$}};
		\node [style=none] (9) at (9.75, 6.25) {\textbf{Candidate sites}};
		\node [style=Cylinder] (11) at (8.75, 0.5) {\rotatebox{270}{$\mathcal{D}_2$}};
		\node [style=Cylinder] (12) at (11, 2.75) {\rotatebox{270}{$\mathcal{D}_k$}};
		\node [style=none] (13) at (-5, 3.25) {};
		\node [style=none] (15) at (4, 3.25) {};
		\node [style=none] (16) at (-0.4, 4.7) {\begin{tabular}{c}
    2. Securely communicate info \\ on $P(\theta| \mathcal{D}_0)$ \end{tabular}};
		\node [style=none] (17) at (4, 2) {};
		\node [style=none] (18) at (-5, 2) {};
		\node [style=none] (20) at (10.25, -1.25) {\begin{tabular}{c}
    3.  Compute $\mathrm{EIG}(e)$ at each site $e$.\end{tabular}};
		\node [style=none] (21) at (-0.25, -1.25) {};
		\node [style=none] (22) at (-10.5, -1) {};
		\node [style=none] (24) at (-0.5, 0.7) {\begin{tabular}{l}
  4.  Return $\{\mathrm{EIG}(e)\}^k_{e=1}$\end{tabular}};
	\end{pgfonlayer}
	\begin{pgfonlayer}{edgelayer}
		\draw [style=Arrow] (13.center) to (15.center);
		\draw [style=Arrow] (17.center) to (18.center); 
	\end{pgfonlayer}
\end{tikzpicture}
    \captionof{figure}{Flow chart depicting our method. In step 1 the \textit{host site} chooses a parameterised model class for the conditional outcome, given by $f_\theta$ and sets the prior $P(\theta)$. 
    They then perform a local Bayesian update, and communicate information on the posterior $P(\theta| \mathcal{D}_0)$ to the \textit{candidate site} using secure multi party computation (MPC). The candidate applies MPC once more to privately calculate the Expected Information Gain (EIG) and communicate it back to host. 
    This allows the host to select the best merge out of the potential candidates.}
    \label{fig:flow_chart}
\end{figure*}

We quantify the value of a merge in a principled, information theoretic way by applying techniques from Bayesian experimental design \citep{rainforth2024modern,lindley1956measure,chaloner1995bayesian}. 
Our solution shares similarities with standard Bayesian dataset acquisition \citep{mackay1992information,kirsch2019batchbald}, however in causal contexts data serve an additional, distinct purpose. 
Specifically acquired data should not only enhance our understanding of the outcome function, but also assist in combating the selection bias that is inherent to causal effect estimation \citep{holland1986statistics} by balancing treatment. 
Put differently, whilst generic data fusion aims at reducing the epistemic uncertainty from incomplete knowledge of the outcome, causal estimation also seeks to improve treatment overlap \citep{rubin1997estimating, crump2009dealing}, i.e. reducing the epistemic uncertainty for counterfactual outcomes.
To resolve this, we make use of favourable parameterisations available in popular Bayesian causal inference methods \citep{hahn2020bayesian,alaa2017bayesian}, which allows us to prioritise information gain in the parameters relevant for the causal problem, rather than gaining information in irrelevant parts of the conditional outcome. 

To ensure privacy, we employ Secure Multi-Party Computation \citep{yao1982protocols,evans2018pragmatic,knott2021crypten}. 
This cryptographic protocol enables multiple parties to jointly compute the output of a function without revealing any of their own private inputs. 
A classic example involves determining the wealthiest person in a group without anyone revealing their personal net worth.
In our context it allows the different candidate sites to compute their expected information gains relative to the initial sites' data, without exposing the contents of their datasets. 
This ensures that the noise required for privacy guarantees can be added to the final statistic as opposed to the raw data at each site. 


Our contributions can be summarised as follows:
\begin{itemize}[leftmargin=18pt]
    \setlength{\itemsep}{8pt} 
    \setlength{\parskip}{-3pt} 
    \item We propose information-theoretic methods to measure the value of a data merge in the context of heterogeneous treatment effects estimation. Our primary contribution is a novel approach that specifically targets the reduction of entropy in parameters that directly influence the conditional average treatment effect (CATE). We also present a standard approach based on expected entropy reduction in all parameters of a conditional outcome model.
    \item We demonstrate how both of the approaches can be used with three popular CATE estimators; Bayesian Polynomial Regression \citep{gelman2021regression}, Causal Multitask Gaussian Processes \citep{alaa2017bayesian}, and Bayesian Causal Forests \citep{hahn2020bayesian}. We derive closed form expressions for the expected entropy reduction in the first two models and give a Monte Carlo estimator for the other.
    \item We provide a privacy protocol for our methods based on multi-party computation \citep[MPC;][]{yao1982protocols}. This ensures that statistic can be computed without any party revealing their raw data. Therefore, differential privacy \citep[DP;][]{dwork2006differential} guarantees can be achieved by noising the \emph{final} computed statistic, rather than to the original raw data, ensuring less loss of accuracy. 
    \item We experimentally validate our methodology across a range of synthetic and semi-synthetic tasks, demonstrating strong agreement between our prospective rankings and the true rankings obtained after performing the merge. Moreover, we find that our proposed methodology to target CATE parameters improves over traditional Bayesian data selection and a number of other baselines. 
    Finally, we show that, for the same level of privacy guarantees, our MPC protocol chained with DP performs better than applying DP to the raw inputs in the linear case. 
\end{itemize}

\section{PROBLEM STATEMENT, ASSUMPTIONS \& NOTATION}\label{sec:background_and_notation}

\paragraph{Notation}
The random variables $X$, $T$, and $Y$ represent the covariates, treatment, and outcomes, with domains $\cX$, $\{0,1\}$, and $\cY$, respectively; $\bx$, $t$, and $y$ denote realisations of these variables.
We let $\cD_e$  be the dataset comprised of elements $\{\bx_i, t_i, y_i \}^{n_e}_{i=1}$, drawn i.i.d. from a distribution $P_{e}(\bx, t, y)$, where $e \in \left\{0,\cdots,K \right\}$ indexes the datasets. 
Vectors of observations in $\cD_e$ are denoted in bold, i.e. $\by_e=(y_i)^{n_e}_{i=1}$, $\bt_e=(t_i)^{n_e}_{i=1}$, and $\bX_e=(\bx_i)^{n_e}_{i=1}$ refers to the data matrix. 
We use potential outcomes framework \citep{rubin1974estimating}, so that $Y(t)$ represents the outcome resulting from an intervention setting $T=t$. 

\paragraph{Assumptions and objective}
Throughout we focus on estimating the 
\emph{Conditional Average Treatment Effect} (CATE), given by:
\begin{align}
    \tau(\bx) = \EE \left[Y(1) -Y(0) | X=\bx \right] \nonumber
\end{align}
To estimate CATE, we begin with an initial dataset, $\cD_0$, referred to as the \textit{host}.
Our goal is to accurately estimate CATE with respect to the distribution of this dataset, $P_0(\bx)$. Specifically, we focus on minimising the Precision in Estimation of Heterogeneous Effects \citep[PEHE;][]{louizos2017causal} given by $\epsilon_{\text{PEHE}}(f)=\int_{P_0(\bx)}\left(\hat{\tau}_f(\bx)-\tau(\bx)\right)^2 d\bx$, where $\hat{\tau}_f$ is the  CATE estimate arising the outcome model $f$. 

We consider a set of potential datasets for merging, $\cD_e$, referred to as the \textit{candidate} sites. 
In these candidate datasets, we assume that the outcomes are unmeasured or masked,
and so denote them by $\cD_{e} = \{\bx^e_i, t^e_i, Y^e_i \}^{n_{e}}_{i=1}$ to show the randomness in $Y^e_i$. 
The goal is to prospectively identify which of the candidate datasets $\cD_e$, would reduce the uncertainty over CATE if we were to measure or unmask the $Y^e_i$'s and merge with the host dataset $\cD_0$.

\begin{figure}
\centering
\scalebox{0.9}{
\begin{tikzpicture}[>=stealth', shorten >=1pt, auto,
    node distance=1.8cm, scale=1.2, 
    transform shape, align=center, 
    state/.style={circle, draw, minimum size=6.0mm, inner sep=0.4mm}]
    \node[state] (v0) at (0,0) {$T$};
    \node[state, above right of=v0] (v2) {$X$};
    \node[state, below right of= v2] (v1) {$Y$};
    \node[state, left of= v2] (v3) {$E$};
    \draw [->, thick] (v0) edge (v1);
    \draw [->, thick] (v2) edge (v0);
    \draw [->, thick] (v2) edge (v1);
    \draw [->, thick] (v3) edge (v2);
    \draw [->, thick] (v3) edge (v0);
    \end{tikzpicture}
    }
   \captionof{figure}{Assumed DAG}
    \label{fig:dag}
\end{figure}
We assume that datasets are generated according to the Directed Acyclic Graph (DAG) in \autoref{fig:dag}.\footnote{We make us of the SWIG framework to combine causal graphical models with potential outcomes. More details can be found in \citet{richardson2013single}.} 
This implies that $P(y | \bx,t, e)$ is fixed across environments, but both covariate distributions and treatment allocation schemes are free to vary i.e.\ $P(\bx,t | e)$ can depend on $e$. 
We also assume positivity over the whole population, so that $\forall~\bx, 0<P(T=1 | X=\bx)<1$, but allow for violations at the site level. 
Adding the consistency assumption (i.e.\ $Y(t) =Y$ when $T=t$) to positivity and the causal structure in \autoref{fig:dag} (which implies no hidden confounders) 
we have that CATE is identifiable~\citep{pearl2009causal,richardson2013single}, constant across environments $e$, and given by:
\begin{align*}
    \tau(\bx)&= \EE \left[Y | X=\bx,T=1 \right] - \EE \left[Y | X=\bx,T=0 \right]\\
    &= \EE \left[Y | X=\bx,T=1,E=e \right]  \\
    & - \EE \left[Y | X=\bx,T=0,E=e \right]
\end{align*}

\section{METHOD}
We take a Bayesian approach to modelling the CATE. Specifically, we define $f_{\theta} : \cX \times \{0,1\} \to \mathcal{Y}$ to be a function representing the expected outcome conditional on covariates and treatment, i.e., $f_{\theta}(\bx, t)$ seeks to approximate $\EE \left[Y | X=\bx,T=t \right]$. We assign a prior distribution $P(\theta)$, to the parameters $\theta$ and denote the conditional likelihood of the outcome given parameters, covariates, and treatment  as $P(Y | \theta, \bx, t)$, which is chosen appropriately based on the type of outcome being modelled.
For example, we can select a normal likelihood for continuous outcomes, or a Bernoulli likelihood for binary ones. 
The data-generating process can be written as
\begin{align*} 
    \theta \sim P(\theta), \quad 
    Y | f_{\theta}(\bx,t) \sim P(Y|\theta,(\bx,t) ). 
\end{align*}
Throughout this paper we focus on continuous $y$ and use a normal likelihood with fixed variance $\sigma^2$, i.e. $P(Y | \bx, t, \theta) = \mathcal{N}(Y; f_{\theta}(\bx, t), \sigma^2)$. CATE is then estimated via the current posterior mean, so that if we have conditioned on data $\cD$ our estimate is $\hat{\tau}(\bx) = \EE_{\theta \sim p(\theta | \mathcal{D})} [f_{\theta}(\bx,1)-f_{\theta}(\bx,0)]$.

\subsection{Quantifying Data Merge Utility through Expected Information Gain}\label{subsec:background_and_cond_outcome}


In order to quantify the value of a data merge, we draw inspiration from Bayesian experimental design \citep[BED;][]{chaloner1995bayesian, rainforth2024modern}. BED applies information theory to provide a measure of what performing a particular experiment would tell us about a parameter of interest, relative to our current beliefs about the parameter value. 
In our context, the `design' of the experiment corresponds to choosing a dataset $\cD_e$, and the outcome is observing  $\{Y^e_{i}\}^{n_e}_{i=1}$.
The value of an experiment is quantified via the expected information gain~\citep[EIG;][]{lindley1956measure}, which measures the expected reduction in uncertainty in the parameters, as measured by Shannon entropy, when moving from the post host posterior, $P(\theta|\cD_0)$ to the post merge posterior, $P(\theta|\cD_0,\cD_e)$: 
\begin{align*}
     \mathrm{EIG}_{\theta | \cD_0}(e) = \EE
     [H[P(\theta | \cD_0)] - H[P(\theta | \cD_{0},\cD_e )]  ],
\end{align*}
where the expectation is over $P(\by_e | \bX_e,\bt_e,\cD_0) = \EE_{P(\theta | \cD_0)}[P(\by_e | \theta, \bX_e,\bt_e)]$---the Bayesian marginal distribution of $\by_e$. 
The EIG is equivalent to the mutual information between parameters and outcomes under the design 
and can be equivalently written as:
\begin{align}
   \mathrm{EIG}_{\theta|\cD_0}(e)  
    = \EE
    \left[ \log \frac{P(\by_e | \theta,\bX_e,\bt_e)}{P(\by_e | \bX_e,\bt_e,\cD_0)}\right],
\label{eq:rearranged_info_gain1}
\end{align}
where the expectation is over $P(\by_e,\theta | \bX_e,\bt_e, \cD_0)$.
This form known as Bayesian Active Learning by Disagreement (BALD) in the active learning literature~\citep{houlsby2011bayesian}. 
For certain classes of models, such as polynomial regression or Gaussian processes, the EIG is available in closed form \citep{sebastiani2000maximum}. In other cases if the likelihood function isn't analytically available, we can approximate it using nested Monte Carlo \citep[NMC;][]{rainforth2018nesting, foster2021variational} as follows:
\begin{align}\label{eq:EIG_estimate_obs}
    \hspace{-7pt}
    \widehat{\mathrm{EIG}}^{\mathrm{NMC}}_{\theta|\cD_0}(e) = \frac{1}{N} \sum^N_{i=1} \log 
    \frac{
        P(\by^{(i)}_e  |\theta^{(i)}, \bX_e,\bt_e)
    }{
       \widehat{P}(\by^{(i)}_{e} | \bX_{e},\bt_{e},\cD_{0}) 
    }, 
\end{align}
%
where $\theta^{(i)}, \by^{(i)}_e \sim P(\theta | \cD_0)P(\by_e | \theta^{(i)},\bX_e,\bt_e)$, and further $M_1$ samples $\theta^{\prime}_j \sim P(\theta | \cD_0)$ for the denominator  
\begin{align}
\hspace{-5pt}
   \widehat{P}(\by^{(i)}_{e} | \bX_{e},\bt_{e},\cD_{0}) = \frac{1}{M_1} \sum^{M_1}_{j=1}P \left(\by^{(i)}_{e} | \theta^{\prime}_j ,\bX_{e},\bt_{e}\right).\label{eq:mc_denom}
\end{align}
Note that since we assume a normal likelihood, we can construct a Rao-Blackwellised estimator by analytically computing the entropy of the likelihood in~\autoref{eq:rearranged_info_gain1} 
to give:
\begin{align*}
\hspace{-3pt}
\widehat{\mathrm{EIG}}_{\theta|\cD_0}^{\mathrm{RB}}(e) = &- \frac{1}{N} \sum^N_{i=1} \log\left( \frac{1}{M_1} \sum_{j=1}^{M_1} P(\by^{(i)}_{e} | \bX_{e},\bt_{e},\theta^{\prime}_j)  \right) \\
&- \frac{n_e}{2}(1+\log(2\pi\sigma^2)).
\end{align*}
%
Nested estimators are biased but consistent, converging to the true EIG at a rate $\cO(({N^{-1} + cM_1^{-2}})^{\frac{1}{2}})$ \citep{rainforth2018nesting}.
A detailed algorithm for both estimators is given in Appendix \ref{ap:EIG_algs}. 

These estimators allow us gauge the value of a merge before observing outcomes $\{Y^e_i\}^{n_e} _{i=1}$ in the dataset $\cD_e$. 
However, whilst this approach provides us with information on the parameters for the conditional outcome, it may not necessarily lead to improved CATE predictions. 
This is because $\mathrm{EIG}_{\theta|\cD_0}$ encourages \emph{uniform entropy reduction} across all dimensions of $\theta$, not just the ones relevant for CATE estimation. For instance, a dataset with untreated individuals only would provide information about the conditional outcome, but less for CATE, which requires viewing treated individuals as well. 
This motivates our improved methodology, which we present in the next section. 

\subsection{EIG Targeting CATE Parameters}\label{subsec:improving_CATE} 
Many causal inference models have additional parameter structure that allows us to target CATE estimation more directly.
Specifically, the parameter set, $\theta$ can often be split as $\theta = \theta_{\text{c}} \cup \theta_{\text{nc}}$ where $\theta_{\text{c}}$ parameterises the CATE model and $\theta_{\text{nc}}$ is a set of nuisance parameters. 
For example, in Bayesian Causal Forests \citep{hahn2020bayesian} the model is parameterised as $f_{\theta}(\mathbf{x},t) = \mu_{\theta_{\text{nc}}}(\mathbf{x}) + t\tau_{\theta_{\text{c}}}(\mathbf{x})$,  
where $\mu_{\theta_{\text{nc}}}$ and $\tau_{\theta_{\text{c}}} (\mathbf{x})$ jointly model the conditional outcome, and $\tau_{\theta_{\text{c}}}(\mathbf{x})$ directly models CATE. 
Leveraging such a parameterisation, we can prioritise uncertainty reduction in $\theta_c$, and therefore CATE, by maximising 
\begin{align}
    \mathrm{EIG}_{\causalparam|\cD_0}(e) &
    = \EE
    \left[ \log  \frac{ P(\by_e | \theta_{\mathrm{c}},\bX_{e},\bt_{e},\cD_{0})}{  P(\by_e | \bX_{e},\bt_{e},\cD_{0}) }  \right],\label{eq:eig_caus_def}
\end{align}
where expectation is over $P(\by_e,\theta_{\mathrm{c}} | \bX_{e},\bt_{e},\cD_{0} )$. Here $P(\by_e | \theta_{\textit{c}},\bX_e,\bt_e,\cD_0) = \EE_{P(\theta_{\textit{nc}} |\theta_{\textit{c}}, \cD_0)}[P(\by_e | \theta, \bX_e,\bt_e)]$ is the Bayesian distribution of the outcomes conditional on the CATE-related parameters only, and is generally not available in closed form. 
To deal with this intractability, we suggest approximating both the numerator and denominator empirically, yielding the following estimator:
\begin{align}
\hspace{-5pt}
   \widehat{\mathrm{EIG}}^{\mathrm{NMC}}_{\causalparam|\cD_0}(e) = \frac{1}{N} \sum^N_{i=1} \log  \frac{
        {\widehat{P}(\by^{(i)}_{e} | \theta^{(i)}_{\mathrm{c}},\bX_{e},\bt_{e},\cD_{0})}
    }{
        \widehat{P}(\by^{(i)}_{e} | \bX_{e},\bt_{e},\cD_{0})
    } \label{eq:EIG_conditional_estimate},
\end{align} 
where $\theta^{(i)}_c, \by^{(i)}_e \sim P(\causalparam | \cD_0)P(\by_e | \causalparam, \bX_e,\bt_e)$, the denominator $\widehat{P}(\by^{(i)}_{e} | \bX_{e},\bt_{e},\cD_{0})$ is as in \autoref{eq:mc_denom}, and use further $M_2$ samples $\theta_{nc}^{(ik)} \sim P(\theta_{nc}^{(ik)} | \theta_{c}^{(i)}, \cD_0)$ for the numerator:
\small
\begin{align}
   {\widehat{P}(\by^{(i)}_{e} | \theta^{(i)}_{\mathrm{c}},\bX_{e},\bt_{e},\cD_{0})}{=}\frac{1}{M_2} \sum^{M_2}_{k=1}P \left(\by^{(i)}_{e} | \theta^{(ik)}_{nc}\cup\theta^{(i)}_c,\bX_{e},\bt_{e} \right) \nonumber
\end{align} 
\normalsize
This ensures that we prioritise a gain in information in the part of the model directly responsible for CATE. We again give an algorithm in Appendix \ref{ap:EIG_algs}. 

\subsection{Procedure and Model Classes}\label{subsec:procedure}\label{sec:model_classes}
We now apply both procedures to three popular Bayesian causal inference methods: Bayesian Polynomial Regression, Bayesian Causal Forests \citep{hahn2020bayesian}, and Causal Multi-task Gaussian Processes \citep{alaa2017bayesian}. We describe the standard predictive method as well as the parameter split used to target CATE.


\paragraph{Bayesian Polynomial Regression} Due to its ubiquity across numerous fields, we first apply our method to Bayesian Polynomial Regression model. This involves specifying an initial polynomial transformation\footnote{This could be an arbitrary non-linear function but we focus on polynomial transformations.} $\phi: \mathcal{X} \times \cT \to \mathbb{R}^p$, a mean $\mu_0$, and precision matrix $\Lambda_0$ to parameterise the prior as: 
\begin{align}
\label{eq:conjugate_linear}
     f_\theta(\bx,t) = \phi(\bx,t)^{\top} \theta, \hspace{1em }\theta \sim \cN( \mu_0, \sigma^2 \Lambda^{-1}_0).
\end{align} 
$\phi$ allows for higher order, or interaction terms. We further assume $\phi(\mathbf{x},t)$ and $\theta$ can be split as:
\begin{align*}
    \phi(\bx,t) = \begin{bmatrix}
           \phi_{\mathrm{nc}}(\bx) & t\phi_{\mathrm{c}}(\bx)
            \\
         \end{bmatrix}^{\top},   \hspace{1em}  {\theta} = \begin{bmatrix}
           \theta_{\mathrm{nc}} &
           \theta_{\mathrm{c}}
           \end{bmatrix}^{\top}
\end{align*}
This covers a broad range of Bayesian polynomial regressions, including the selection used in~\citet{gelman2021regression}. 
In Appendix \ref{ap:proof_bayes_lin} we show that both EIGs can be computed in closed form:
\begin{proposition}
For the Bayesian Polynomial Regression model defined in \autoref{eq:conjugate_linear} we have: 
\begin{align*}
 \hspace{-2pt}   
    \mathrm{EIG}_{\theta | \cD_0}(e) &{=} \log 
    {\det \left(\Phi_e^{\top}\Phi_e 
    + \Phi_0^{\top}\Phi_0+\Lambda_0 \right)}
    {+} C \\ 
    \hspace{-2pt}   
    \mathrm{EIG}_{\theta_{\mathrm{c}} | \cD_0}(e) &{=} \log 
    \det(\Phi_{c,e}^{\top} \Phi_{c,e} + \Phi_{c,0}^{\top}\Phi_{c,0}+\Lambda_0^{[\mathrm{c},\mathrm{c}]})
    {+} C^{\prime},
\end{align*}
where $\Phi_e=\phi(\bX_e,\bt_e)$, $\Phi_0=\phi(\bX_0,\bt_0)$, $\Phi_{c,e}=\bt_e \odot\phi_{\mathrm{c}}(\bX_e)$, $\Phi_{c,0}=\bt_0\odot\phi_{\mathrm{c}}(\bX_0)$, with $\phi,\phi_c$ applied row-wise and $\odot$ denoting element-wise multiplication; $C,C^{\prime}$ are constant in $e$. 
\end{proposition}

\paragraph{Bayesian Causal Forest} Bayesian Causal Forests \citep[BCF;][]{hahn2020bayesian} are one of the most popular causal inference methods, building upon Bayesian Additive Regression Trees \citep[BART;][]{chipman2012bart}, which are themselves a mainstay in observational causal inference \citep{hill2011bayesian}. 
The BCF model can be expressed as $f_{\theta}(\mathbf{x},t) = \mu_{\theta_{\text{nc}}}(\mathbf{x}) + t\tau_{\theta_{\text{c}}}(\mathbf{x}),$ 
%
where $\mu_{\theta_{\text{nc}}} $ and $\tau_{\theta_{\text{c}}}(\mathbf{x})$ are independent BART models; 
further details and alternative parameterisations can be found in Appendix \ref{ap:model_via_sample}. 
As the posterior is only available via sampling, we estimate both EIGs using NMC as given in \autoref{eq:EIG_estimate_obs} and \autoref{eq:EIG_conditional_estimate}.

\paragraph{Causal Multi-task Gaussian Processes} Causal multi-task Gaussian Processes \citep{alaa2017bayesian} use a vector-valued GP \citep{alvarez2012kernels} to jointly model the conditional outcomes, allowing information sharing between them: 
%
\begin{align}
    \mathbf{f} = \begin{bmatrix}
           f(\bx,0) &
           f(\bx,1)
         \end{bmatrix}^{\top}, \text{ where } \mathbf{f} \sim \mathcal{GP}(0,\bK) \label{eq:mtgp_model}
\end{align}
for a vector-valued kernel $\bK: \cX \times \cX \to \mathbb{R}^{2 \times 2}$. The outcomes are then modelled by evaluating the relevant portion of the GP. Under this setting CATE is given by $\Tilde{\tau} = \mathbf{e} \mathbf{f}$ for $\mathbf{e} =\begin{bmatrix}  -1 & 1        \end{bmatrix}$, meaning that $\Tilde{\tau}$ is also a GP given by $\Tilde{\tau} \sim \mathcal{GP}(0,\mathbf{e}^{\top}\bK\mathbf{e})$. The advantage of causal multi-task GPs is that they allow us to get a closed form posterior for $\tau$ without having to observe any samples from $Y(1)-Y(0)$. 

As GPs are inherently non-parametric they do not fit directly into the framework laid out in \autoref{subsec:background_and_cond_outcome} and \autoref{subsec:improving_CATE}. 
This is not a problem for the predictive case as we can replace $\theta$ with $\mathbf{f}$ in \autoref{eq:rearranged_info_gain1} and get closed form expressions \citep{houlsby2011bayesian}. 
However, for the causal case this creates challenges as we cannot directly evaluate the expressions in \autoref{eq:eig_caus_def} with $\Tilde{\tau}$ in the place of $\theta_{\textit{c}}$. To resolve this we instead focus on entropy reduction in CATE predictions on the host dataset. 
We denote this by $\Tilde{\tau}(\bX_0)$, where $\bX_0$ is the host data matrix. 
The information gains $e$ denote these by $\mathrm{EIG}_{\mathbf{f}|\cD_0}(e)$ and $\mathrm{EIG}_{\Tilde{\tau}(\bX_0)|\cD_0}(e)$ respectively. 
As the following proposition shows, both of these are now available in closed form:

\begin{proposition}
Let $n_e^{(t)}$ be the number of subjects receiving treatment $t$ in dataset $e$. For the causal multi-task GP model, defined in \autoref{eq:mtgp_model} we have 
\begin{align*}
\mathrm{EIG}_{\mathbf{f}|\cD_0} &= \frac{1}{2}\log \det \left( \Sigma_1\right) \\
&- n^{(0)}_e\log(\sigma_0) - n^{(1)}_e\log(\sigma_1) \\
    \mathrm{EIG}_{\Tilde{\tau}(\bX_0)|\cD_0}(e) &= \frac{1}{2}\log \left({\det (\Sigma_1) \det(\Sigma_2) } \right) \\
    &- \frac{1}{2}\log \left({\det(\Sigma) }\right),
\end{align*}
where $\Sigma_1,\Sigma_2,\Sigma$ and the proof are given Appendix~\ref{ap:CMGP}.
\end{proposition}

\section{PRIVACY}\label{sec:privacy}

For privacy we use \textit{Multi-Party Computation}~\citep[MPC;][]{evans2018pragmatic}. 
First introduced by \citet{yao1982protocols}, MPC focuses on a setting where $m$ separate parties wish to compute the value of a function $f(x_1,\cdots,x_m)$ where the $i^{\mathrm{th}}$ party inputs $x_i$ and wishes to keep this private. To resolve this, MPC involves the specification of a protocol of message passing between parties which if followed would lead to the computation of $f$. In this work, we focus on the \emph{semi-honest} setting, in which all parties follow the specified protocol, but some \emph{corrupt} parties will try to learn as much about their peers inputs in the process. The goal is to devise a protocol which will preserve the privacy of the non-corrupt party's inputs, up to a given computational budget by the adversary. In our setting this means that any collection of corrupt sites are unable to learn anything about the other sites data during the EIG calculation, so any noise needed for privacy guarantees can be added to the final statistic. 

For implementing multi-party computation, we employ the open source library CrypTen \citep{knott2021crypten}. CrypTen builds upon PyTorch \citep{paszke2019pytorch} allowing for standard tensor operations to be performed in an MPC protocol. 
For arithmetic operations on floating-point values this is achieved as follows: A float, $x_F$, is multiplied by some large scaling factor $B=2^{L}$ and rounded to the nearest integer $\lfloor x_F \rceil$, where $L$ is the number of precision bits. 
The integer $\lfloor x_F \rceil$ is then associated with its equivalence class $x \in \mathbb{Z}/Q\mathbb{Z}$ where $\mathbb{Z}/Q\mathbb{Z}$ is a ring of $Q$ elements. The value $x$ can then be shared across all $m$ parties using Shamir secret sharing \citep{shamir1979share}, in which each party gets access to a share of $x$ is given by $\left[ x \right]_i \in \mathbb{Z}/Q\mathbb{Z}$ which is generated such that the sum of all shares recovers the original value, so $x = \sum^m_{i=1} \left[ x \right]_i\mod Q$. At any point all parties can combine their shares to decode the output as  $x_F \approx x/B$. We let $[x] = \{ \left[ x \right]_i\}^m_{i=1}$ denote the set of all shares corresponding to the secret value $x$.

Arithmetic operations building on addition are performed locally, so that for two secret values $[x],[y]$ each party performs $[z]_i = [x]_i+[y]_i$ and the result, $z$ is obtained by all parties summing their share. M
ultiplication is implemented using Beaver triples \citep{beaver1992efficient}, logarithms are approximated using householder iterations \citep{householder1970numerical}, and reciprocals use Newton-Raphson.
We implement log-determinants using Cholesky LDL decompositions, which are preferred to standard Cholesky factorisations as they avoid the use of square roots, which would require additional approximation in Crypten. 
This is possible as we only compute the log determinant of positive semi-definite matrices. 
This provides all the operations necessary to implement the above $\mathrm{EIG}$ calculations in a private manner using MPC. 

When returning EIG statistics to the host, we add a small amount of noise to prevent information leakage.
If we only need to output the best site, we use the exponential protocol  \citep{dwork2006differential} ensuring minimal information leakage. 
Finally, we note that the discretisation required to represent a float in the ring $\mathbb{Z}/Q\mathbb{Z}$ involves some degree of precision loss. 
Nevertheless, as we empirically demonstrate in the next section, this leads to minimal depreciation in performance compared to differential privacy.


\section{EXPERIMENTS \& RESULTS}
\begin{figure}[t]
    \begin{center}
      \includegraphics[width=0.47\textwidth]{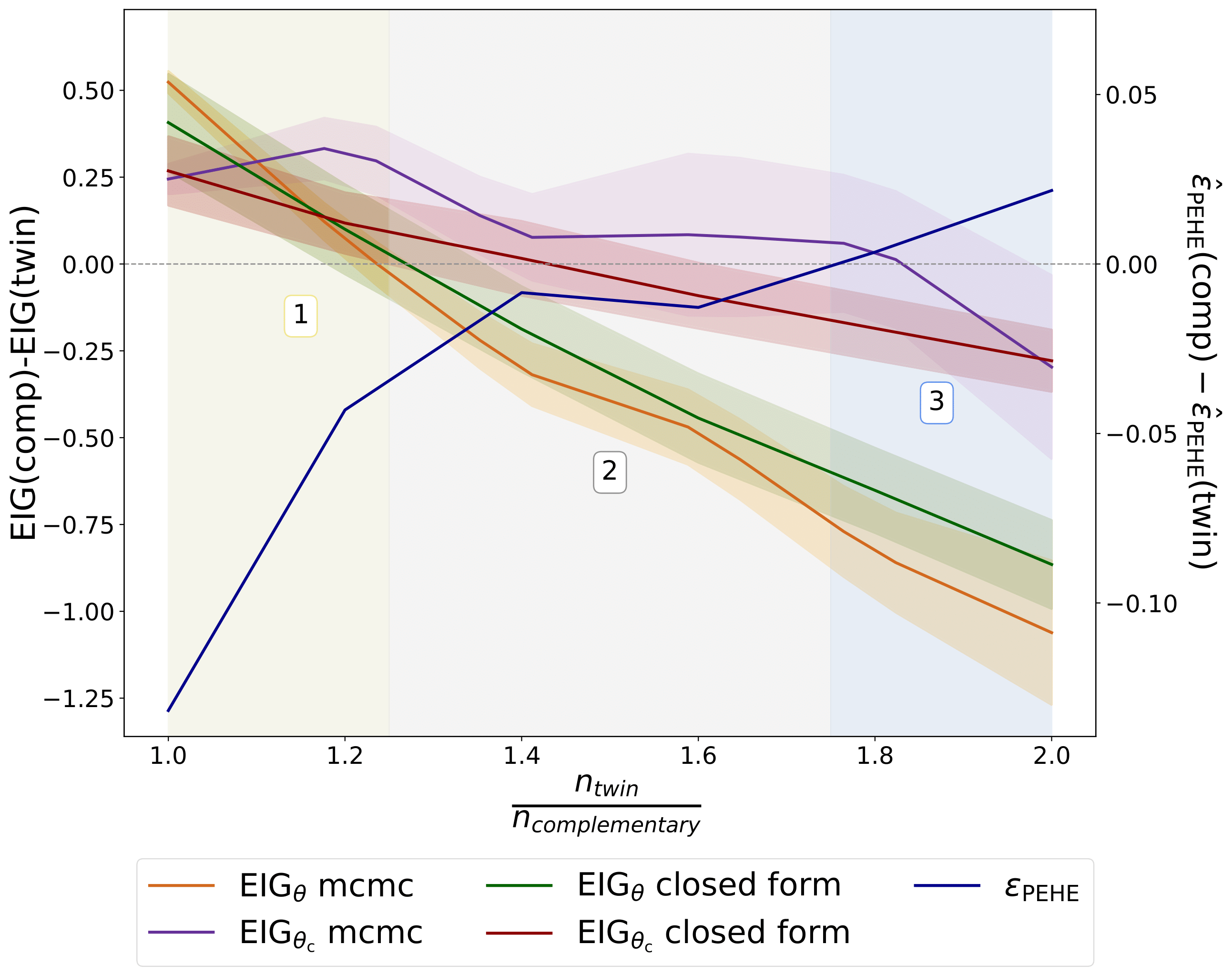}
    \end{center}
    \caption{
    Difference in post host $\mathrm{EIG}$  and $\hat{\epsilon}_{\text{PEHE}}$ for a linear CATE model trained on $\cD_0\cup\cD_{\textit{comp}}$ and $\cD_0\cup\cD_{\textit{twin}}$ for increasing $\frac{n_{\textit{twin}}}{n_{\textit{comp}}}$ and fixed $n_{\textit{host}} {=} n_{\textit{comp}} {=} 100$. PEHE is evaluated on hold out data from the host distribution. Lines show mean$\,\pm1\,$s.d. 
    across $50$ seeds. The three regions show different datasets preferences. 
    In region 2, $\mathrm{EIG}_{\theta|\cD_0}$ incorrectly favours $\cD_\textit{twin}$, whilst $\mathrm{EIG}_{\theta_c|\cD_0}$ correctly selects  $\cD_\textit{comp}$ for the causal task. 
    }
    \label{fig:linexp_synthetic}
\end{figure} 
We experimentally validate our approach in a setting where the host has to rank a number of candidate datasets based on the estimated gain from merging. We use selection of synthetic and semi-synthetic benchmark datasets as this allows us to use the known CATE to get a ground truth ranking. For each model, we do this by ranking datasets on the true PEHE on $P_0(\bx)$ of the relevant model trained on the merged dataset $\cD_0 \cup \cD_e$. This is then compared against implied EIG rankings. Throughout, we tune any model hyper-parameters on the host site when measuring the information gain as well as re-tuning paramters on each merged dataset when getting the ground truth ranking.

\begin{table*}
    \centering
    \begin{tabular}{llcccc}
        \toprule
        \textbf{Model} & \textbf{Objective} & $\boldsymbol{\rho} (\uparrow)$ & \textbf{p@1} $(\uparrow)$ & \textbf{p@3} $(\uparrow)$ & \textbf{p@5} $(\uparrow)$ \\ 
        \midrule
        \multirow{3}{*}{Polynomial} 
            & $\mathrm{EIG}_{\theta_{\textit{c}|\cD_0 }}$ & $\mathbf{0.70 \pm 0.08}$  & $\mathbf{0.50 \pm 0.15}$ & $\mathbf{0.70 \pm 0.04}$ & $\mathbf{0.78 \pm 0.04}$  \\
            & $\mathrm{EIG}_{\theta|\cD_0}$            & $\mathbf{0.68 \pm 0.06}$  & $\mathbf{0.50 \pm 0.15}$ & $\mathbf{0.70 \pm 0.06}$ & $\mathbf{0.76 \pm 0.04}$  \\
            & Best baseline                           & ${0.40 \pm 0.11}$         & ${0.40 \pm 0.15}$         & ${0.60 \pm 0.15}$         & ${0.66 \pm 0.04}$         \\
        \midrule
        \multirow{3}{*}{Causal GP} 
            & $\mathrm{EIG}_{\Tilde{\tau}(X_0)|\cD_0 }$  & $\mathbf{0.49 \pm 0.06}$  & $\mathbf{0.50 \pm 0.15}$ & $\mathbf{0.50 \pm 0.08}$ & $\mathbf{0.62 \pm 0.06}$  \\
            & $\mathrm{EIG}_{\mathbf{f}|\cD_0}$           & ${0.33 \pm 0.06}$         & ${0.30 \pm 0.15}$         & ${0.43 \pm 0.05}$         & $\mathbf{0.60 \pm 0.04}$  \\
            & Best baseline                                & ${0.31 \pm 0.12}$         & ${0.10 \pm 0.20}$         & ${0.20 \pm 0.07}$         & ${0.46 \pm 0.05}$         \\
        \midrule
        \multirow{3}{*}{Bayesian CF} 
            & $\mathrm{EIG}_{\theta_{\textit{c}|\cD_0 }}$  & $\mathbf{0.54 \pm 0.10}$  & $\mathbf{0.60 \pm 0.15}$ & $\mathbf{0.63 \pm 0.08}$ & $\mathbf{0.70 \pm 0.04}$  \\
            & $\mathrm{EIG}_{\theta|\cD_0}$              & $0.36 \pm 0.10$           & $0.30 \pm 0.14$           & $0.50 \pm 0.07$           & $\mathbf{0.66 \pm 0.05}$  \\
            & Best baseline                           & $0.45 \pm 0.11$           & $\mathbf{0.60 \pm 0.14}$    & $\mathbf{0.63 \pm 0.08}$    & $\mathbf{0.70 \pm 0.04}$  \\
        \bottomrule
    \end{tabular}
    \caption{Ranking experiment for the IHDP dataset, measured by Spearman $\rho$ and precision at k (p@k). We include the best performing baseline method, which is different for different models.  }
    \label{tab:random_exp}
\end{table*}

To generate the datasets, we begin with an initial, large dataset $\cD$ from which we subsample the host and candidate sites. We do this by choosing a selection function, $S_e(\bx,t)$, and using it to subsample a dataset of size $n_e$, where $S_e(\bx_i,t_i)$ is the probability of subsampling point $i$ for dataset $\cD_e$. Varying the selection functions across sites ensures heterogeneity amongst datasets whilst complying with the independences given by the causal structure in \autoref{fig:dag}. We begin by providing an illustrative example on synthetic data before evaluating on the causal benchmarks, specifically: Lalonde \citep{lalonde1986evaluating} and Infant Health and Development Program \citep[IHDP;][]{louizos2017causal}. Further details and results can be found in Appendix \ref{ap:exp_detail} and \ref{ap:further_exp}, respectively.

\subsection{Illustrative experiment}
\paragraph{Concept} To illustrate the difference between our two methods---the standard approach based on the full parameter set, $\mathrm{EIG}_{\theta}$ (\autoref{eq:rearranged_info_gain1}), and the CATE-targeting one, $\mathrm{EIG}_{\theta_c}$ (\autoref{eq:eig_caus_def})---we start with a simple example where the host must choose between two candidate sites.
We design these sites as follows:
a \emph{complementary} site, representing the ``ideal'' merge for causal purposes,  and a \emph{twin} site, containing information similar to the host. 
To create these, we first simulate an initial large dataset $\cD$ as if it were a randomised controlled trial with an equal probability of treatment and subsample the host dataset, $\cD_0$, using a selection function $S_0(\bx, t)$. 
The data of the complementary site, $\cD_{\emph{comp}}$, is subsampled using $1-S_0(\bx,t)$ as a selection function, whilst the twin site uses $S_0(\bx,t)$. This ensures the twin dataset mirrors the host's distribution and the complementary causally ``complements'' it. 
Assuming equal sizes, merging $\cD_{0}$ with $\cD_{\textit{comp}}$ would recreate the initial randomised trial $\cD$, as the variable $e$ acts as a collider for $\bx$ and $t$, 
removing their conditional dependency (see causal DAG in Appendix \ref{fig:dag_illustrative_exp}).
Therefore, $\cD_\textit{comp}$ represents an ideal merge as it balances treatment allocation, whereas $\cD_{\textit{twin}}$ covers similar regions of the data space to those in the host, $\cD_0$, potentially amplifying pre-existing biases and imbalances.

For the illustrative experiment, we vary the ratio of sample sizes, $\frac{n_{\textit{twin}}}{n_{\textit{comp}}}$, in order to compare which dataset is chosen by $\mathrm{EIG}_{\theta|\cD_0}$ and the causally targeted $\mathrm{EIG}_{\causalparam|\cD_0}$ for linear regression. The aim is to demonstrate that $\mathrm{EIG}_{\theta|\cD_0}$ will prefer $\cD_{\textit{twin}}$ at points where $\cD_{\textit{comp}}$ dataset is still the preferable dataset for CATE estimation. Indeed, for large values of the ratio $\frac{n_{\textit{twin}}}{n_{\textit{comp}}}$, $\cD_{\textit{twin}}$ provides significant information about the conditional outcome, but not in the regions that are most relevant for causal estimation. On the other hand, $\mathrm{EIG}_{\causalparam}$ should continue to select $\cD_{\textit{twin}}$ whilst it remains preferable for CATE estimation. We simulate $\bx \in \mathbb{R}^3$ where $x_1$ is Bernoulli and other covariates are normal. The true outcome is sampled from a normal linear model, and the selection functions are logistic regressions. We also include sample based estimates for each $\mathrm{EIG}$ to demonstrate how they differ from their closed form counterparts. Experimental details provided in Appendix \ref{ap:exp_detail}.

\paragraph{Results} \autoref{fig:linexp_synthetic} shows the results of the experiment divided into three regions. In region one, both methods choose the complementary dataset over the twin dataset which is consistent with ground truth ranking given by the PEHE upon merging. In region two, $\mathrm{EIG}_{\theta|\cD_0}$ chooses $\cD_{\textit{twin}}$ whilst $\mathrm{EIG}_{\theta_{\textit{c}|\cD_0 }}$ opts for $\cD_{\textit{comp}}$. Here, the complementary dataset is still the optimal in terms of CATE, but $\mathrm{EIG}_{\theta|\cD_0}$ preferences $\cD_{\textit{twin}}$ as it leads to a greater entropy reduction in the full set of parameters. This result shows that by focusing on the causal parameters alone, $\mathrm{EIG}_{\theta_{\textit{c}|\cD_0 }}$ is able to make the correct decision in selecting $\cD_{\textit{comp}}$. In the final region we see all lines have crossed the $x$ axis, showing that all methods agree with the ground truth in choosing $\cD_{\textit{twin}}$. Finally, we note the MCMC estimates agree with the closed form counterparts, with increased variability due to sampling.

\subsection{Ranking experiment}\label{sec:ranking}

\paragraph{Concept} For our main experiment we validate our framework in a setting where the host must choose between  many potential candidates, each with different distributions. 
To do so we begin with a standard causal inference benchmark dataset, $\cD$, and form the host and candidates datasets using subsampling functions, $S_e(\bx,t)$ as detailed above where subsampling function is a logistic regression with random parameters. This ensures that each site has a different covariate distribution. We apply the three methods described in Section \autoref{sec:model_classes} to estimate both the two expected information gains for each candidate site, $\cD_e$. Ultimately, like before, we compare the implied rankings with the ground truth ranking given by the PEHE. Full details are provided in Appendix~\ref{ap:exp_detail}.

\paragraph{Baselines} 
We provide a number of simple comparison methods as baselines for our task. Specifically, (a) ranking by sample size, (b) ranking by similarity of covariate distribution measured by a multivariate Gaussian fit to the host, and (c) ranking by dissimilarity of treatment allocation measured by the error of a propensity model fit on the host. We compare using Spearman rho($\rho$) and precision at $k$ (p@$k$).

\paragraph{Results}
\autoref{tab:random_exp} shows the results of our experiment on the IHDP dataset \citep{louizos2017causal} where the host has to rank the best datasets out of 10 candidates. We report the average performance of the rankings across 20 repeated experiments, and according to Spearman $\rho$ \citep{spearman1961proof} and precision at k. We include the best baseline by Spearman $\rho$ performance. Additional metrics and baseline performance can be found in Appendix \ref{ap:further_exp}. These results demonstrate that across all models, our $\mathrm{EIG}_{\causalparam}$ based approach, focusing on causal parameters, outperforms the standard approach which seeks expected entropy reduction in all parameters. Further, our method works significantly better than all baselines.
\begin{table}
\label{tab:mpc_exp}
\vspace{0pt}
    \begin{minipage}{\linewidth}
        \centering
        \resizebox{\linewidth}{!}{\begin{tabular}{lrr}
        \toprule
        \textbf{Method}     & \textbf{MSE} $(\downarrow)$  & $\boldsymbol{\rho}(\uparrow)$  \\
        \midrule
        \textbf{MPC Linear}      & $\mathbf{(3.80 \pm 0.04) \times 10^{-6}}$   & $\mathbf{0.797 \pm 0.06}$ \\  
        \textbf{DP Linear}        & $9.80 \pm 0.30$  & ${0.06} \pm 0.11$  \\
        \bottomrule
        \end{tabular}%
        }
\caption{Multi-Party Computation Results for $\mathrm{EIG}_{\causalparam}$.} 
\label{tab:mpc_exp}
\end{minipage}
\end{table}
%
\subsection{Multi-Party Computation Experiments}\label{sec:MPC_experi}
Finally, we experimentally demonstrate the performance of our cryptographic protocol. We repeat a similar experiment as in \autoref{sec:ranking} on the IHDP dataset, this time choosing between twenty datasets. Results are given using a linear model as this allows us to compare our MPC based approach against differential privacy 
\citep[DP;][see Appendix \ref{ap:privacy} for definition]{dwork2006differential}. We use $\epsilon=100$ and Laplace noising \citep{dwork2014algorithmic}, following existing work on DP linear regression \citep{bernstein2019differentially,awan2021structure}.  
In order to ensure a fair comparison, we noise the final statistics from the MPC computation with an appropriate amount of Laplace noise. This comes from the sensitivity of the EIG statistic which we derive in Appendix \ref{ap:sens_linear_stat}. \autoref{tab:mpc_exp} shows both the MSE induced by the privacy method and the Spearman $\rho$ against the noise-free ranking. Results show that MPC vastly outperforms differential privacy, both in terms of MSE and ranking, with DP producing a near random ranking, despite the relaxed noising. 



\section{RELATED WORK}

\paragraph{Federated Learning Via Multi-Party Computation} Our setting bares large with federated learning \citep{li2020review}, a distributed machine learning approach that enables multiple parties to collaboratively train a shared model while keeping their raw data decentralised and private. Our goal differs from this as we focus not on learning a model across sites, but deciding which sites to pool. The application of multiparty computation within federated learning is a growing area \citep{li2020privacy,mugunthan2019smpai,kanagavelu2020two}, with much of the work focusing on how to learn predictive models in a secure cross site fashion. 
Most similar to our work is \citet{muazu2024federated}, who develop a similar  federated approach to data fusion focusing on healthcare, however they do not take a causal approach. To the best of our knowledge, there are no existing applications of multi-party computation within causal inference.

\paragraph{Federated/Private Causal Estimation} There are, however, federated approaches to estimating causal effects. In this area, a majority of works partition the loss function into multiple components, with each component corresponding to a specific data source \citep{vo2022adaptive, liu2024disentangle, vo2023federated}. However, modelling complex, non-linear relationships remains challenging \citep{almodovar2023federated}. 
Many of these algorithms come without privacy guarantees, with the exception of \citet{niu2022differentially}, who add DP guarantees to various popular CATE estimation techniques. 
The latter approach however contrasts with ours as the use of sample splitting reduces data efficiency. 

\paragraph{Causal Bayesian Active Learning} 
Bayesian Active Learning by Disagreement \citep[BALD;][]{houlsby2011bayesian} is a framework for strategic training data acquisition, focusing on regions of high uncertainty. Our method based on maximising $\mathrm{EIG}_{\theta|\cD_0}$  (\autoref{eq:rearranged_info_gain1}) can be viewed as applying BALD after an initial update to an entire datasets rather than individual datapoints. 
Most similar to our work is CausalBALD \citep{jesson2021causal}, which applies an active learning approach to CATE estimation. However, the acquisition function cannot easily be extended to datasets without ignoring the correlation in information provided by different points. Most applications of active learning in causality focus on causal discovery and intervention selection \citep{toth2022active,hauser2014two,annadani2023differentiable}.

\section{DISCUSSION}

\paragraph{Limitations} Due to the cost of computing high dimensional determinants and performing multiple rounds of conditional sampling, our method can be computationally costly in high dimensions. Moreover, the cost of multi-party computation will also increase as higher order approximations are required to retain accuracy. However, both of these remain negligible compared to the data engineering and expenses of data fusion \citep{brodie2010data, kadadi2014challenges}. 
Finally, whilst our method offers a secure and principled way to prospectively quantify the value of dataset merges, the amount of information needed to justify such expenses may vary depending on application. It might, therefore, be beneficial to consider introducing a problem-specific threshold to determine when a proposed merger is worthwhile.

\paragraph{Conclusion}
We introduce an information-theoretic, cryptograpically secure framework for evaluating the utility of potential data merges for causal estimation. To the best of our knowledge, this is the first work addressing this relevant challenge. Through empirical evaluation, we demonstrate that our framework can reliably rank datasets according to their ability to improve CATE estimation. We show that entropy reduction in the CATE parameters alone gives an improvement when compared to a more standard approach to Bayesian dataset acquisition. Finally, we demonstrate that our cryptographic procedure can be applied in conjunction with DP, resulting in lower loss of accuracy, compared to applying DP alone.

\section*{Acknowledgements}
 We would like to thank Andrew Yiu, Amartya Sanyal, Shahine Bouhabid, Xi Lin, Patrick Hough, and Sahra Ghalebikesabi for their comments. JF gratefully acknowledges funding from the EPSRC. LTM is supported by EPSRC through the Modern Statistics and Statistical Machine Learning (StatML) CDT programme, grant no. EP/S023151/1.

\bibliographystyle{abbrvnat} 
\bibliography{ref}

\newpage
\section*{Checklist}



 \begin{enumerate}

 \item For all models and algorithms presented, check if you include:
 \begin{enumerate}
   \item A clear description of the mathematical setting, assumptions, algorithm, and/or model. [Yes]
   \item An analysis of the properties and complexity (time, space, sample size) of any algorithm. [Yes]
   \item (Optional) Anonymized source code, with specification of all dependencies, including external libraries. [Yes]
 \end{enumerate}

 \item For any theoretical claim, check if you include:
 \begin{enumerate}
   \item Statements of the full set of assumptions of all theoretical results. [Yes]
   \item Complete proofs of all theoretical results. [Yes]
   \item Clear explanations of any assumptions. [Yes]     
 \end{enumerate}

 \item For all figures and tables that present empirical results, check if you include:
 \begin{enumerate}
   \item The code, data, and instructions needed to reproduce the main experimental results (either in the supplemental material or as a URL). [Yes]
   \item All the training details (e.g., data splits, hyperparameters, how they were chosen). [Yes]
         \item A clear definition of the specific measure or statistics and error bars (e.g., with respect to the random seed after running experiments multiple times). [Yes]
         \item A description of the computing infrastructure used. (e.g., type of GPUs, internal cluster, or cloud provider). [Yes]
 \end{enumerate}

 \item If you are using existing assets (e.g., code, data, models) or curating/releasing new assets, check if you include:
 \begin{enumerate}
   \item Citations of the creator If your work uses existing assets. [Yes]
   \item The license information of the assets, if applicable. [Not Applicable]
   \item New assets either in the supplemental material or as a URL, if applicable. [Not Applicable]
   \item Information about consent from data providers/curators. [Not Applicable]
   \item Discussion of sensible content if applicable, e.g., personally identifiable information or offensive content. [Not Applicable]
 \end{enumerate}

 \item If you used crowdsourcing or conducted research with human subjects, check if you include:
 \begin{enumerate}
   \item The full text of instructions given to participants and screenshots. Not Applicable]
   \item Descriptions of potential participant risks, with links to Institutional Review Board (IRB) approvals if applicable. [Not Applicable]
   \item The estimated hourly wage paid to participants and the total amount spent on participant compensation. [Not Applicable]
 \end{enumerate}

 \end{enumerate}

\appendix
\onecolumn
\section{Mathematical Details}
\vspace{1em}
\subsection{Algorithms for Computing EIG}\label{ap:EIG_algs}
\vspace{1em}

\begin{algorithm}[ht]
\hspace*{\algorithmicindent} \textbf{Input}: Data Matrix $\bX_e$, Treatment Vector $\bt_e$, Variance $\sigma^2$ \\
 \hspace*{\algorithmicindent} \textbf{Output}: $\widehat{\mathrm{EIG}}^{\mathrm{NMC}}_{\theta|\cD_0}(e)$
\caption{Algorithm for $\widehat{\mathrm{EIG}}^{\mathrm{NMC}}_{\theta|\cD_0}(e)$}\label{alg:cap}
\begin{algorithmic}
\Require $l = N M_1$ for some $n,M_1>0$
\State $S \gets 0$
\State Sample $\{\theta_{i} \}^l_{i=1} \sim P(\theta \mid \cD_{0})$ for $l=N M_1$
\State Split $\{\theta_{i} \}^l_{i=1}$ as $\{(\theta_{i}),(\theta_{i,j})^{M_1}_{j=1} \}^N_{i=1}$
 \For{$i \in \{1,\cdots,N\}$}
 \State Sample $\by^{(i)}_e \sim P(\by_e| \bX_e,\bt_e,\theta_i)$
\State $S \gets S +\log 
    \frac{P(\by^{(i)}_e  |\theta_{i}, \bX_e,\bt_e)}{\frac{1}{{M_1}} \sum^{M_1}_{j=1}P(\by^{(i)}_e | \theta_{i,j}, \bX_e,\bt_e) }$
  \EndFor
\State $\widehat{\mathrm{EIG}}^{\mathrm{NMC}}_{\theta|\cD_0}(e) \gets  \frac{1}{N} S$
\State \Return $\widehat{\mathrm{EIG}}^{\mathrm{NMC}}_{\theta|\cD_0}(e)$
\end{algorithmic}
\end{algorithm}

\vspace{0.5em}

\begin{algorithm}[ht]
\hspace*{\algorithmicindent} \textbf{Input}:$\{\theta_{i} \}^l_{i=1} \sim P(\theta \mid \cD_{0})$, Data Matrix $\bX_e$, Treatment Vector $\bt_e$, Variance $\sigma^2$ \\
 \hspace*{\algorithmicindent} \textbf{Output}: $\widehat{\mathrm{EIG}}^{\mathrm{RB}}_{\theta|\cD_0}(e)$
\caption{Algorithm for $\widehat{\mathrm{EIG}}^{\mathrm{RB}}_{\theta|\cD_0}(e)$}\label{alg:cap}
\begin{algorithmic}
\State $S \gets 0$
\State Sample $\{\theta_{i} \}^l_{i=1} \sim P(\theta \mid \cD_{0})$ for $l=N M_1$
\State Split $\{\theta_{i} \}^l_{i=1}$ as $\{(\theta_{i}),(\theta_{i,j})^{M_1}_{j=1} \}^N_{i=1}$
 \For{$i \in \{1,\cdots,N\}$}
 \State Sample $\by^{(i)}_e \sim P(\by_e| \bX_e,\bt_e,\theta_i)$
\State $S \gets S - \log 
    {\frac{1}{{M_1}} \sum^{M_1}_{j=1}P(\by^{(i)}_e | \theta_{i,j}, \bX_e,\bt_e) }$
  \EndFor
\State $\widehat{\mathrm{EIG}}^{\mathrm{RB}}_{\theta|\cD_0}(e) \gets  \frac{1}{N} S$
\State \Return $\widehat{\mathrm{EIG}}^{\mathrm{NMC}}_{\theta|\cD_0}(e)$
\end{algorithmic}
\end{algorithm}

\vspace{0.5em}

\begin{algorithm}[H]
 \hspace*{\algorithmicindent} \textbf{Input} Data Matrix $\bX_e$, Treatment Vector $\bt_e$, Variance $\sigma^2$ \\
 \hspace*{\algorithmicindent} \textbf{Output}: $\widehat{\mathrm{EIG}}_{\causalparam|\cD_0}(e)$
\caption{Algorithm for $\widehat{\mathrm{EIG}}_{\causalparam|\cD_0}(e)$}\label{alg:cap}
\begin{algorithmic}
\State $S \gets 0$  
\State Sample $\{\theta_{i} \}^N_{i=1} \sim P(\theta \mid \cD_{0})$
 \For{$i \in \{1,\cdots,N\}$}
 \State Sample $ \by_e  \sim P(\by_e|\theta,\bX_e,\bt_e)$
  \State Sample $\{\theta^{\prime}_j \}^{M_1}_{j=1} \sim P(\theta \mid  \cD_{0})$
 \State Sample $\{(\theta_{\mathrm{nc}})^{(ik)} \}^{M_2}_{k=1} \sim P(\theta_\mathrm{nc} \mid (\causalparam)_i, \cD_{0})$
\State $\widehat{P}(\by^{(i)}_{e} | \bX_{e},\bt_{e},\cD_{0}) \gets \frac{1}{M_1} \sum^{M_1}_{j=1}P \left(\by^{(i)}_{e} | \theta^{\prime}_j ,\bX_{e},\bt_{e}\right)$
\State ${\widehat{P}(\by^{(i)}_{e} | \theta^{(i)}_{\mathrm{c}},\bX_{e},\bt_{e},\cD_{0})} \gets\frac{1}{M_2} \sum^{M_2}_{k=1}P \left(\by^{(i)}_{e} | \theta^{(ik)}_{nc}\cup\theta^{(i)}_c,\bX_{e},\bt_{e} \right)$
\State $S \gets S + \log\left(  \frac{{\widehat{P}(\by^{(i)}_{e} | \theta^{(i)}_{\mathrm{c}},\bX_{e},\bt_{e},\cD_{0})}}{\widehat{P}(\by^{(i)}_{e} | \bX_{e},\bt_{e},\cD_{0})} \right)$
  \EndFor
\State \Return $\frac{1}{N} S$
\end{algorithmic}
\end{algorithm}

\subsection{Differential Privacy Definition}\label{ap:privacy}

\begin{definition}
 We say a randomised algorithm, $\cA$ satisfies $\epsilon$ differential privacy if for any input dataset $\cD$ and dataset $\cD^{\prime}$ differing by a single entry, we have
 \begin{align*}
     P( \cA(\cD) \in \cO) \leq \exp(\epsilon)P( \cA(\cD^{\prime}) \in \cO).
 \end{align*}
\end{definition}

\subsection{Sensitivity of Linear Statistic}\label{ap:sens_linear_stat}

\vspace{1em}
We derive the sensitivity of the linear EIG statistic in order to give a fair comparison with naive differential privacy in \autoref{sec:MPC_experi}.
\begin{proposition}
    Let:
\begin{align}
    f(\bX_e) = \log\det(\bX^{\top}_e\bX_e +\bX^{\top}_0\bX_0+ \Lambda_0)
\end{align}
If $\Lambda_0 = cI$ and $\Norm{\bx}_{\infty}\leq M$ for all $\bx \sim P_e(\bx)$ and $e$.  Then we have that:
\begin{align}
    \Delta_f = \max_{\bX^{\prime}_e,\bX_e}\norm{f(\bX^{\prime}_e) - f(\bX_e)} \leq \frac{Md}{\sqrt{c}}
\end{align}
Where $\bX^{\prime}_e,\bX_e$ differ in at most one row. This implies $f(\bX_e)+Z$ for $Z \sim \mathrm{Laplace}(\frac{Md}{\epsilon\sqrt{c}})$ is a $\epsilon$ differentially private release of $f(\bX_e)$. 
\end{proposition}
\begin{proof}
     To prove this we will use the fact that if $\max_{\Norm{\bX}\leq M}\Norm{Df(\bX)}_{F} =  L$ we have that
    $\norm{f(\bX^{\prime}_e) - f(\bX_e)} \leq L \Norm{\bX^{\prime}_e - \bX_e}_{F}$
for all $\bX^{\prime}_e, \bX_e$ with norm bounded by $M$ where $\Norm{\cdot}_{F}$ is the Frobenious norm. Write $\bE = \bX^{\top}_0\bX_0+ \Lambda_0$, via the chain rule we have that $Df(\bX) = \circbrac{\bX_e^{\top}\bX_e+\bE}^{-1}\bX_e^{\top}$
Now:
\begin{align}
\Norm{\circbrac{\bX_e^{\top}\bX_e+\bE}^{-1}\bX_e^{\top}}_{F} &= \sqrt{\mathrm{tr}\circbrac{\circbrac{\bX_e^{\top}\bX_e+\bE}^{-1} \bX_e^{\top}\bX_e \circbrac{\bX_e^{\top}\bX_e+\bE}^{-1} }} \\
&= \sqrt{\mathrm{tr}\circbrac{\circbrac{\bX_e^{\top}\bX_e+\bE}^{-1} - \circbrac{\bX_e^{\top}\bX_e+\bE}^{-1} \bE \circbrac{\bX_e^{\top}\bX_e+\bE}^{-1}}}\\
&\leq \sqrt{\mathrm{tr}\circbrac{\circbrac{\bX_e^{\top}\bX_e+\bE}^{-1} }} \leq \sqrt{\frac{d}{c}}
\end{align}
We have used the fact that $\circbrac{\bX_e^{\top}\bX_e+\bE}^{-1} \bE \circbrac{\bX_e^{\top}\bX_e+\bE}^{-1}$ is positive semi definite and so has positive trace, and that the eigenvalues of $\bX_e^{\top}\bX_e+\bE$ are bounded below by $c$  so the eigenvalues of $\circbrac{\bX_e^{\top}\bX_e+\bE}^{-1}$ are bounded above by $\frac{1}{c}$. Finally for neighbouring datasets we can change at most $d$ entries by $M$ so $\Norm{\bX^{\prime}_e - \bX_e}_{F} \leq \sqrt{d}M$
\end{proof}

\section{Model Details}\label{ap:bayes_causal_models}
Throughout all methods will aim to model the outcome as some function plus an error, so:
\begin{align}
    Y_i = f(\bx_i,t_i)+ \epsilon_i
\end{align}
\subsection{Models via Sampling}\label{ap:model_via_sample}
\paragraph{Bayesian Additive Regression Trees (BART)}
BART models \citep{chipman2012bart}  $f$ as the sum of $L$ piecewise constant binary regression trees, so we have:
\begin{align}
    f(\bx,t) = \sum^L_{l=1} g_i(\bx,t,T_l,\bm_l)
\end{align}
where $T_l$ is a regression tree given by a partition $(\cA_1,\cdots,\cA_{\cB(l)})$ of $\cX \times \cT$ and the set of leaf parameter values $\bm_l=(m_{l1},\cdots,m_{l\cB(l)})$ so that:
\begin{align}
    g_l(\bx,t) = m_j \text{ if } \bx,t \in \cA_j
\end{align}
The mean parameters are given with independent normal parameters $m_{lj}\sim \cN(0,\sigma_m)$. Over trees, the prior is such that the probability of a node having children at depth $d$ is given by:
\begin{align}
    \alpha(1+d)^{-\theta} \text{ for } \alpha \in (0,1), \theta \in [0,\infty)
\end{align}
The original BART model explores this space using Metropolis-Hastings Markov chain Monte Carlo, but we make use of XBART \citep{he2019xbart} for accelerated posterior sampling. 
\paragraph{Bayesian Causal Forest} 
Bayesian Causal Forests \citep[BCF;][]{hahn2020bayesian} build upon BART models utilising specific parameterisations for causal inference tasks. 
The two parameterisations suggested in \citet{hahn2020bayesian} are firstly:
\begin{align}
    f(\mathbf{x},t) = \mu(\mathbf{x}) + t\tau(\mathbf{x})
\end{align}
Where $\mu,\tau$ are independent BART models. To draw specific attention to their parameters will write them as $\mu_{\theta_{\text{nc}}},\tau_{\theta_{\text{c}}}$ noting that $\tau_{\theta_{\text{c}}}$ is a model for CATE. 
\citet{hahn2020bayesian} note that this parameterisation is not invariant to which treatment is assigned as positive or negative, leading them to propose the following invariant parameterisation:
\begin{align}
    f_{\theta}(\mathbf{x},t) = \Tilde{\mu}_{\Tilde{\theta}_{\text{nc}}}(\mathbf{x}) + b_t\Tilde{\tau}_{\Tilde{\theta}_{\text{c}}}(\mathbf{x})
\end{align}
Where $b_t \sim \cN(0,\frac{1}{2})$. Under this parameterisation a CATE estimate is given by $(b_1-b_0)\Tilde{\tau}_{\Tilde{\theta}_{\text{c}}}(\mathbf{x})$.
When sampling we make use of the accelerated BCF approach \citep{krantsevich2023stochastic} which builds upon XBART and uses the following slightly modified model:
\begin{align}
    f_{\theta}(\mathbf{x},t) = a\Tilde{\mu}_{\Tilde{\theta}_{\text{nc}}}(\mathbf{x}) + b_t\Tilde{\tau}_{\Tilde{\theta}_{\text{c}}}(\mathbf{x})
\end{align}
Where $a\sim\cN(0,1)$. 

We define the set $\causalparam$ to be any parameters affiliated with the $\tau$ model, including $b_t$ for the invariant parameterisation. In order to sample $P(\theta_{\text{nc}} | \causalparam,\cD_0)$ we refit $\theta_{\text{nc}}$ parameters on the dataset $\cD_0$ to the residuals resulting from subtracting the $\tau$ portion of the model. So this refitting $\mu$ is as follows for the standard parametisation:
\begin{align}
    Y -t{\tau}_{{\theta}_{\text{c}}}(\mathbf{x}) = {\mu}_{{\theta}_{\text{nc}}}(\mathbf{x}) 
\end{align}
Or for the accelerated BCF approach \citep{krantsevich2023stochastic}:
\begin{align}
    Y- b_t\Tilde{\tau}_{\Tilde{\theta}_{\text{c}}}(\mathbf{x}) = a\Tilde{\mu}_{\Tilde{\theta}_{\text{nc}}}(\mathbf{x}) 
\end{align}
Where any parameters on the left hand side are fixed. 
\subsection{Closed Form Models}
In this section we give details of models for which the EIG is available in closed form. We provide details of the models as well as proofs for the expressions. 

\subsubsection{Bayesian Polynomial Regression Derivations}\label{ap:proof_bayes_lin}
In this we derive the results for Bayesian polynomial regression. We have modelled our data as:
\begin{align}
     y \sim \cN ( \phi(\bx,t)^{\top} \theta ,\sigma^2), \hspace{1em }\theta \sim \cN( \mu_0, \sigma^2 \Lambda^{-1}_0)
\end{align}
In this context, the posterior is available in closed form as:
\begin{align}
    \theta | \cD_{0} &\sim \cN(\mu_{0},\sigma^2\Tilde{\Lambda}^{-1}_{0}) \\
    \Tilde{\Lambda}_{0} &= \left(\phi \left(\bX_{0},\bt_{0}\right)^{\top}\phi \left(\bX_{0},\bt_{0}\right) + \Lambda_0^{-1} \right) \\
    \mu_{0} &= \left( \Lambda_{0}\right)^{-1} \left( \phi \left(\bX_{0},\bt_{0}\right)^{\top}\phi \left(\bX_{0},\bt_{0}\right) \hat{\theta}_{0} + \Lambda_0 \mu_0 \right) \\
    \hat{\theta}_{0} &= \left(\phi \left(\bX_{0},\bt_{0}\right)^{\top}\phi \left(\bX_{0},\bt_{0}\right) \right)^{-1} \phi \left(\bX_{0},\bt_{0}\right)^{\top}\by_{0}
\end{align}

\paragraph{Expected Information Gain Over all parameters} We use the fact that $\mathrm{EIG}_{\theta|\cD_0}$ can be written as:
\begin{align*}
     \mathrm{EIG}_{\theta | \cD_0}(e) = \EE_{P(\by_e | \bX_e,\bt_e,\cD_0)} \left[ H[P(\theta | \cD_0)] - H[P(\theta | \cD_{0},\cD_e )]  \right],
\end{align*}
As the posterior over $\theta$ is Gaussian we can directly evaluate these expressions using the closed form entropy for a Gaussian distribution as:
\begin{align*}
    H[P(\theta | \cD_0)] = \frac{n_e}{2} \left(1+ \log(2\pi)\right)+\frac{1}{2}\left( \log \det \left( \Tilde{\Lambda}^{-1}_{0} \right)\right)
\end{align*}
The distribution $\theta | \cD_{0},\cD_e )$ can be obtained as above, where we now use $\Tilde{\Lambda}_{0}$ as the prior precision matrix before updating on $\cD_{0}$. This gives:
\begin{align*}
    H[P(\theta | \cD_0)] = \frac{n_e}{2} \left(1+ \log(2\pi)\right)+\frac{1}{2}\left( \log \det \left( \left(  \phi \left(\bX_{e},\bt_{e}\right)^{\top}\phi \left(\bX_{e},\bt_{e}\right) + \Tilde{\Lambda}_{0} \right)^{-1}  \right)\right)
\end{align*}
Using the above form for $\Tilde{\Lambda}_{0}$ and collecting all constants gives the expression presented in the text.

\paragraph{ $\mathrm{EIG}_{\theta_{\textit{c}}|\cD_0}$: Expected Information Gain Over all parameters} This follows as above but using the fact that the block form of the matrices to allow us write the covariance precision matrix for the post host posterior over $\theta_{\textit{c}}$ as follows:
\begin{align*}
\left(\bt_0 \odot\phi_{\mathrm{c}}(\bX_0)\right)^{\top}\left(\bt_0 \odot\phi_{\mathrm{c}}(\bX_0)\right)+(\Lambda_0)_{[\mathrm{c},\mathrm{c}]}
\end{align*}
Where $(\Lambda_0)_{[\mathrm{c},\mathrm{c}]}$ corresponds to block of $\Lambda_0$ with entries after $[c,c]$ in the row and column. 
\subsection{Causal Multitask Gaussian Processes \citep{alaa2017bayesian}} \label{ap:CMGP}
In this work, CATE is modelled using a multitask Gaussian process \citep{bonilla2007multi}. Multitask Gaussian Processes use a GP in vector-valued Reproducing Kernel Hilbert Space (vv-RKHS) to share information between tasks \citep{alvarez2012kernels}. In \citet{alaa2017bayesian}, learning the conditional outcome function for each treatment is seen as a separate task, so we jointly model:
\begin{align}
    Y | \mathbf{x},t \sim \cN(0,f_t({\bx}),\sigma_t^2) 
\end{align}
Where each $f_t$ is a Gaussian Process. The kernel $\bK_{\eta}:\cX \times \cX \to \mathbb{R}^{2 \times 2}$ is now a symmetric
positive semi-definite matrix-valued function, with hyper-parameters $\eta$. In the case of \citet{alaa2017bayesian} they use a \emph{linear model of coregionalization}\footnote{See \citet{alvarez2012kernels} for more details.}, giving the kernel as:
\begin{align}
    \bK_{\eta}(\bx,\bx^{\prime}) = \bA_0 k_0(\bx,\bx^{\prime}) + \bA_1 k_1(\bx,\bx^{\prime})
\end{align}
Where $k_{t}$ is the RBF kernel, given by:
\begin{align}
    k_t(\bx,\bx^{\prime}) = \exp(-\frac{1}{2} \left(\bx-\bx^{\prime} \right)^{\top}\bR^{-1}_t\left(\bx-\bx^{\prime} \right))
\end{align}
Where $\bR^{-1}_t = \mathrm{diag}\left( \ell^2_{1,t},\cdots,\ell^2_{d,t} \right)$ and $\ell_{j,t}$ is the length-scale parameter for the treatment $T=t$ in the $j^{\mathrm{th}}$ coordinate of $\bx$. $\bA_t$ is given by:
\begin{align}
    \mathbf{A}_0=\left[\begin{array}{cc}
\theta_{00}^2 & \rho_0 \\
\rho_0 & \theta_{01}^2
\end{array}\right], \mathbf{A}_1=\left[\begin{array}{cc}
\theta_{10}^2 & \rho_1 \\
\rho_1 & \theta_{11}^2
\end{array}\right].
\end{align}
Where $\theta_ij$ and $\rho_i$ determine the variances and covariances of the shared tasks $f_t$. So we have that the full set of hyper-parameters $\eta = (\theta_0,\theta_1,\bR_0,\bR_1,\bA_0,\bA_1)$. Once all these hyper-parameters have been learnt we have that the covariance between different function evaluations, $f_t(\bx),f_{t^{\prime}}(\bx^{\prime})$, is given the $t,t^{\prime}$ coordinate of $\bK_{\eta}(\bx,\bx^{\prime})$. So:
\begin{align}
    \mathrm{cov}(f_t(\bx),f_{t^{\prime}}(\bx^{\prime})) = \bK_{\eta}(\bx,\bx^{\prime})_{[t,t^{\prime}]}
\end{align}
Now, if we let $K((\bx,t),(\bx^{\prime},t^{\prime}) = \bK_{\eta}(\bx,\bx^{\prime})_{[t,t^{\prime}]}$ then we can obtain the posterior kernel in a similar way to the standard case. Precisely if we the training data be given by:
\begin{align}
    \Tilde{\mathbf{X}}&=\left[\left\{\bx_i\right\}_{T_i=0},\left\{\bx_i\right\}_{T_i=1}\right]^T, \\
    \Tilde{\mathbf{Y}}&=\left[\left\{y_i^{\left(T_i\right)}\right\}_{T_i=0},\left\{y_i^{\left(T_i\right)}\right\}_{t_i=1}\right]^T, \\
\boldsymbol{\Sigma}&=\operatorname{diag}\left(\sigma_0^2 \mathbf{I}_{n-n_1}, \sigma_1^2 \mathbf{I}_{n_1}\right) \\
n_1&=\sum_i W_i, \\
\mathbf{K}_\eta(x)&=\left(\mathbf{K}_\eta\left(x, X_i\right)_i.\right)
\end{align}
Then we have that the posterior mutlitask GP has mean and posterior kernel given by:
\begin{align}
& m^{\mathrm{post}}(\bx) = \mathbf{K}_{\eta}^T(\bx) \left(\bK_{\eta}(\bX,\bX)+\Sigma \right)^{-1} \Tilde{\mathbf{Y}} \\
& \mathbf{K}^{\mathrm{post}}_{\eta^*}(\bx, \bx^{\prime}) = \mathbf{K}_{\eta^*}(\bx, \bx^{\prime})-\mathbf{K}_{\eta^*}(\bx) \left(\bK_{\eta}(\bX,\bX)+\Sigma \right)^{-1} \mathbf{K}_{\eta^*}^T(\bx^{\prime}) \\
\end{align}
This leads to the following posterior over CATE, where $\mathbf{e} = [-1,1]^{\top}$:
\begin{align}
    \Tilde{\tau}(x) \sim \cN(m^{\mathrm{post}}(\bx)^{\top}\mathbf{e}, \mathbf{e}^{\top} \mathbf{K}^{\mathrm{post}}_{\eta^*}(\bx, \bx^{\prime})\mathbf{e} )
\end{align}
\subsubsection{Expected Information Gain}\label{ap:CMGP_proof} 

First, let $\mathbf{X}^{(1)}_e,\mathbf{X}^{(0)}_e$ and $\mathbf{y}^{(1)}_e,\mathbf{y}^{(0)}_e$ be the covariance and outcomes for environment $e$ that is treated and untreated respectively. To avoid confusion with treatment we will use $\bX_{e^*}$ to refer to the host environment for this derivation. We will also use $\bK_{|\cD_0}$ to refer to the posterior kernel.   Now to derive the Expected Information Gain in closed form we need the distribution of the following vector:
\begin{align}
    \begin{bmatrix}
           \mathbf{y}^{(1)}_e \\
           \mathbf{y}^{(0)}_e \\
           \Tilde{\tau}(\bX_{e^*}) 
           \end{bmatrix} | \bX_e,\cD_0\sim \cN \left( \bm, \Sigma  \right)
\end{align}
Where we have that:
\begin{align}
     &\Sigma_1 = \begin{bmatrix}
           \bK_{|\cD_0}(\mathbf{X}^{(0)}_e,\mathbf{X}^{(0)}_e)+ \sigma_0^2 I_{n^{0}_e} &  \bK_{|\cD_0}(\mathbf{X}^{(1)}_e,\mathbf{X}^{(0)}_e)) \\
           \bK_{|\cD_0}(\mathbf{X}^{(0)}_e,\mathbf{X}^{(1)}_e) & \bK_{|\cD_0}(\mathbf{X}^{(1)}_e,\mathbf{X}^{(1)}_e)+ \sigma_1^2 I_{n^{1}_e} \\
           \end{bmatrix} \\
&\Sigma_2 = \bK_{|\cD_0}(\mathbf{X}^{(1)}_{e^*},\mathbf{X}^{(1)}_{e^*})+\bK_{|\cD_0}(\mathbf{X}^{(0)}_{e^*},\mathbf{X}^{(0)}_{e^*}) - 2\bK_{|\cD_0}(\mathbf{X}^{(1)}_{e^*},\mathbf{X}^{(0)}_{e^*}) \\
&\Sigma =\begin{bmatrix}
            \Sigma_1 & \Sigma_{12} \\
            \Sigma_{12}^{\top} & \Sigma_2
           \end{bmatrix} \text{ where } \Sigma_{12} = \begin{bmatrix}
\bK_{|\cD_0}(\mathbf{X}^{(1)}_{e^*},\mathbf{X}^{(0)}_e)-\bK_{|\cD_0}(\mathbf{X}^{(0)}_{e^*},\mathbf{X}^{(0)}_e) \\
\bK_{|\cD_0}(\mathbf{X}^{(1)}_{e^*},\mathbf{X}^{(1)}_e)-\bK_{|\cD_0}(\mathbf{X}^{(0}_{e^*},\mathbf{X}^{(1)}_e) 
\end{bmatrix}
\end{align}
Now from the covariance matrix we can derive the standard Expected Information Gain $\mathrm{EIG}_{\theta|\cD_0}$ and CATE-specific $\mathrm{EIG}_{\theta_{\textit{c}|\cD_0 }}$. Throughout we will use $\bK$ to the posterior kernel irrespective if it has been fit or not.
\paragraph{
Expected Information Gain over the conditional outcome parameters}
For the $\mathrm{EIG}_{\mathbf{f}}$ we use the BALD form:
\begin{align}
    H(\by_e | \bX_e,\cD_0) - H(\by | \bX_e,\mathbf{f}, \cD_0)
\end{align}
Using the standard form of entropy for a Gaussian distribution we can read $H(\by_e | \bX_e,\cD_0)$ off of the covariance matrix above as:
\begin{align}
   H(\by_e | \bX_e,\cD_0) &= \frac{n_e}{2} \left(1+ \log(2\pi)\right)+\frac{1}{2}\left( \log \lvert \left( \Sigma_1\rvert \right)\right) \\
   \text{where }\Sigma_1 &= \begin{bmatrix}
            \bK_{\eta}(\mathbf{X}^{(0)}_e,\mathbf{X}^{(0)}_e)+ \sigma_0^2 I_{n^{0}_e} &  \bK_{\eta}(\mathbf{X}^{(1)}_e,\mathbf{X}^{(0)}_e)) \\
           \bK_{\eta}(\mathbf{X}^{(0)}_e,\mathbf{X}^{(1)}_e) & \bK_{\eta}(\mathbf{X}^{(1)}_e,\mathbf{X}^{(1)}_e)+ \sigma_1^2 I_{n^{1}_e} \\
           \end{bmatrix}
\end{align}
And $H(\by_e | \mathbf{f},\bX_e,\cD_0)$ being:
\begin{align}
    H(\by_e | \mathbf{f},\bX_e,\cD_0) = \frac{n_e}{2} \left(1+ \log(2\pi)\right)+\frac{1}{2} \left(n^{(0)}_e\log(\sigma^2_0) +n^{(1)}_e\log(\sigma^2_1)\right)
\end{align}
This gives the Expected Information Gain as:
\begin{align}
    \cI_{\mathbf{f}}(e) = \frac{1}{2}\log \lvert \left( \Sigma_1\rvert \right) - \frac{1}{2}\left(n^{(0)}_e\log(\sigma^2_0) +n^{(1)}_e\log(\sigma^2_1) \right)
\end{align}

\paragraph{
Expected Information Gain on the CATE parameters}
We now target an Expected Information Gain on the CATE parameters on our host dataset, so $\Tilde{\tau}(\bX_0)$. By using the fact that the Expected Information Gain can be written as the mutual information between $\Tilde{\tau}(X_0)$ and the observed outcomes in dataset $e$, we use the closed form mutual information for Gaussian's to directly write this as:
\begin{align}
    \cI_{\Tilde{\tau}(\bX_0)}(e) &= \frac{1}{2}\log \left(\frac{\lvert \Sigma_1\rvert \lvert \Sigma_2 \rvert }{\lvert \Sigma\rvert } \right) \\
    \text{where }\Sigma_2 &= \bK_{\eta}(\mathbf{X}^{(1)}_{e^*},\mathbf{X}^{(1)}_{e^*})+\bK_{\eta}(\mathbf{X}^{(0)}_{e^*},\mathbf{X}^{(0)}_{e^*}) - 2\bK_{\eta}(\mathbf{X}^{(1)}_{e^*},\mathbf{X}^{(0)}_{e^*})
\end{align}

\section{Experimental Details}\label{ap:exp_detail}

\begin{figure}[t]
\centering
\begin{subfigure}[t]{.5\textwidth}
  \scalebox{0.8}{
\begin{tikzpicture}[>=stealth', shorten >=1pt, auto,
    node distance=1.8cm, scale=1.2, 
    transform shape, align=center, 
    state/.style={circle, draw, minimum size=6.0mm, inner sep=0.4mm}]
    \node[state] (v0) at (0,0) {$T$};
    \node[state, above right of=v0] (v2) {$X$};
    \node[state, below right of= v2] (v1) {$Y$};
    \node[state, left of= v2] (v3) {$E$};
    \draw [->, thick] (v0) edge (v1);
    \draw [->, thick] (v2) edge (v1);
    \draw [->, thick] (v2) edge (v3);
    \draw [->, thick] (v0) edge (v3);
    \end{tikzpicture}
    }
    \caption{Before merging: causal structure in $D_{\textit{host}}$, or $D_{\textit{twin}}$ or $D_{\textit{comp}}$ \textit{taken separately}. E acts as a collider and thus creates a dependency between $X$ and $T$}
\end{subfigure}%
\begin{subfigure}[t]{.5\textwidth}
  \centering
  \scalebox{0.8}{
    \begin{tikzpicture}[>=stealth', shorten >=1pt, auto,
    node distance=1.8cm, scale=1.2, 
    transform shape, align=center, 
    state/.style={circle, draw, minimum size=6.0mm, inner sep=0.4mm}]
    \node[state] (v0) at (0,0) {$T$};
    \node[state, above right of=v0] (v2) {$X$};
    \node[state, below right of= v2] (v1) {$Y$};
    \draw [->, thick] (v0) edge (v1);
    \draw [->, thick] (v2) edge (v1);
    \end{tikzpicture}
    }
    \caption{After merging: causal structure in $D_{\textit{host}} \cup D_{\textit{comp}}$}
\end{subfigure}
\caption{Causal structure for the illustrative experiment.}
\label{fig:dag_illustrative_exp}
\end{figure}

\paragraph{General experimental settings and hyperparameters} 

All standard deviations and precisions were taken equal to 1. Throughout experiments, priors were taken as zero-valued vector. 

In the \textbf{illustrative experiment}, 400 samples were used for computing outer expectations, and 800 samples for inner expectations. Here, we consider $X = (X_0, X_1, X_2) \in \mathbb{R}^3$. We use the sampling selection function $P_{\text{host}}(x,t)= \text{sigmoid}(1 + 2 \times x_0 - x_1 + 2 \times t) + \epsilon$ and outcome model $y = 1 + x_0 - x_1 + x_2 + 5 \times t + 2 \times x_0 + 2 \times x_0 - 4 \times x_2 + \epsilon$ with $X_0 \sim \mathcal{B}(12, 3)$, $X_1 \sim \mathcal{N}(4,1)$, $X_2 \sim \mathcal{B}(1, 7)$ and $\epsilon\sim \cN (0,1)$. 

In both \textbf{ranking experiments}, the selection functions are randomly generated. We first generate a binary vector of the size of the dimension of $X$ to define the subset of covariates that would be impact selection. We then generate two other random vectors, one for the multiplicative coefficients for each selected covariate, and another to define a power for each term in the sum.  Ultimately, the probability of selection is taken as the sigmoid of this randomly generated polynomial.

In the \textbf{ranking experiment with IHDP}, 10 candidates were generated with a sample size ranging from 300 to 500. The host sample size was equal to 400. The experiment was across 20 seeds. We kept a minimum of 50 subjects in each treatment group. The hold out test dataset  had a sample size of 2000. For the linear model, $X$, $T$ and $X \times T$ were included. For the Gaussian Process model, a maximum of 1000 iterations was set. In CBF, both the predictive and conditional models were used with a maximum depth of 250, and a shrinkage $\alpha=0.95$. 

In the \textbf{ranking experiment with Lalonde}, 15 candidates were generated with a sample size ranging from 200 to 400. The host sample size was equal to 600. The experiment was across 20 seeds. We kept a minimum of 50 subjects in each treatment group. The hold out test dataset had a sample size of 2000. For the linear model, $X$, $T$ and $X \times T$ were included. For the Gaussian Process model, a maximum of 1000 iterations was set. In CBF, both the predictive and conditional models were used with a maximum depth of 200, and a shrinkage $\alpha=0.9$.

\paragraph{Datasets} We describe the two datasets used in our experiments, with high-level summary given in \autoref{tab:app_dataset_description}.

\begin{table}[h]
\vspace{5pt}
\caption{Description of the datasets: Lalonde \citep{lalonde1986evaluating} and IHDP \citep{louizos2017causal}.} \label{tab:app_dataset_description}
    \centering
    \begin{tabular}{lcc}
    \toprule
    \textbf{}           & \textbf{ihdp} & \textbf{lalonde} \\ \midrule
    \textbf{Number of samples}  &    747        &     16,177             \\
    \textbf{Number of features}   &        24      &       8           \\ \bottomrule
    \end{tabular}%
\vspace{10pt}
\end{table}

The \textbf{Infant Health and Development Program, or IHDP} is a randomized controlled study designed to assess how home visits by specialist doctors impact the cognitive test scores of premature infants. Initially, the dataset serves as a benchmark for evaluating treatment effect estimation algorithms, as described in \cite{hill2011bayesian}. This evaluation introduces selection bias by excluding non-random subsets of treated individuals to construct an observational dataset, with outcomes derived from the original covariates and treatments. 

The \textbf{Lalonde} originates from the National Supported Work Demonstration used by  \citet{dehejia1999causal} to evaluate propensity score matching methods. The data consists of demographic variables (age, race, academic background, and previous real earnings), as well as a treatment indicator. The outcome is the real earnings in the year 1978.


\paragraph{Compute times}
Approximate compute times for the ranking experiment on causal benchmark datasets are given in \autoref{tab:approx_compute_times_appendix}. 
Experiments were performed on  an Apple M3 chip with a 12-core CPU and 18 GB of RAM.
\begin{table}[H]
\centering
\caption{Approximate compute times. 
} \label{tab:approx_compute_times_appendix}
\begin{tabular}{lcc}
\toprule
\textbf{}           & \textbf{ihdp} & \textbf{lalonde} \\ \midrule
\textbf{Polynomial}    &    <  1 min        &     < 1 min            \\
\textbf{Causal GP}     &         3 mins      &       3 mins           \\
\textbf{BART}       &       14 mins       &       9 mins                       \\ \bottomrule
\end{tabular}%
\end{table}

\vspace{3em}
\begin{figure*}[ht]
    \centering
    \begin{subfigure}[t]{0.45\textwidth}
        \centering
        \includegraphics[width = 0.8\textwidth]{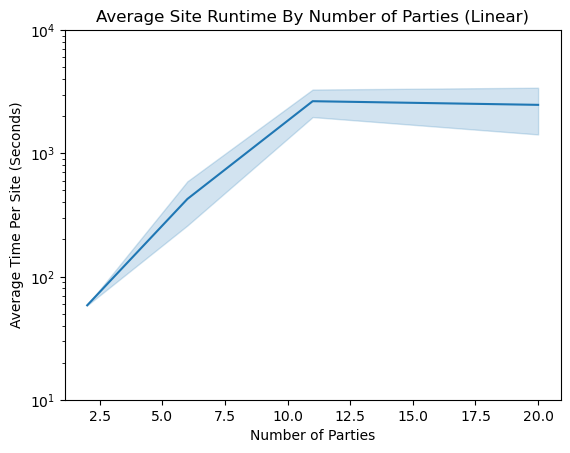}
        \caption{In this we repeat the multi-party computation experiments from \autoref{sec:MPC_experi} varying the number of sites. Each site is treated as a separate party in the multi-party computation and the average runtime per site is reported. Runtime recorded on an M3 macbook. Averaged over 5 runs.}
    \end{subfigure}%
    ~ 
    \hspace{1em}
    \begin{subfigure}[t]{0.45\textwidth}
        \centering
        \includegraphics[width = 0.8\textwidth]{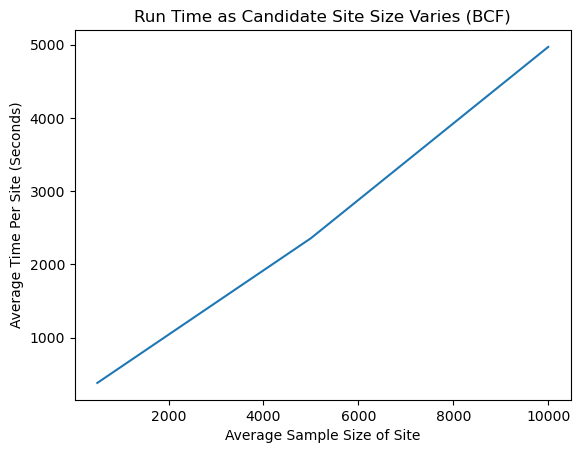}
        \caption{Runtime of Bayesian Causal Forest as the candidate sample size increases. We run the algorithm for the task of selecting between 5 sites with average size 500, 5000, and 10,000.  }
    \end{subfigure}%
\end{figure*}
\section{Further experimental results}\label{ap:further_exp}

For completeness, we include the performance of all baselines on the IHDP dataset in \autoref{tab:IHDP_results} and the Lalonde in \autoref{tab:Lalonde_results}.

\begin{table}[t]
\centering
\caption{IHDP dataset ranking experiment results with 10 candidate datasets} \label{tab:IHDP_results}
\begin{tabular}{llcccc}
        \toprule
                \textbf{Model} & \textbf{Objective} & $\boldsymbol{\rho} (\uparrow)$ & \textbf{p@1} $(\uparrow)$ & \textbf{p@3} $(\uparrow)$ & \textbf{p@5} $(\uparrow)$ \\ \midrule
        \multirow{5}{*}{Polynomial} &
        $\mathrm{EIG}_{\theta_{\textit{c}|\cD_0 }}$ & $\mathbf{0.70 \pm 0.08}$  & $\mathbf{0.50 \pm 0.15}$ & $\mathbf{0.70 \pm 0.04}$ & $\mathbf{0.78 \pm 0.04} $  \\
        & $\mathrm{EIG}_{\theta|\cD_0}$  & $\mathbf{0.68 \pm 0.06}$ & $\mathbf{0.50 \pm 0.15}$ & $\mathbf{0.70 \pm 0.06}$ & $\mathbf{0.76 \pm 0.04}$  \\
        & PropScore Error  & ${0.40 \pm 0.11}$  & ${0.40 \pm 0.15}$ & ${0.60 \pm 0.15}$  & ${0.66 \pm 0.04}$  \\
        & Sample Size  & ${0.34 \pm 0.08}$  & ${0.10 \pm 0.08}$ & ${0.27 \pm 0.04}$  & ${0.50 \pm 0.06}$  \\
        & CovDist  & ${0.03 \pm 0.07}$  & ${0.10 \pm 0.08}$ & ${0.23 \pm 0.06}$  & ${0.48 \pm 0.04}$  \\
        \midrule
        \multirow{5}{*}{Causal GP} & 
         $\mathrm{EIG}_{\Tilde{\tau}(X_0)|\cD_0 }$         & $\mathbf{0.49 \pm 0.06}$ & $\mathbf{0.50 \pm 0.15}$ & $\mathbf{0.50 \pm 0.08}$ & $\mathbf{0.62 \pm 0.06}$ \\
        & $\mathrm{EIG}_{\mathbf{f}|\cD_0}$        & ${0.33 \pm 0.06}$ & ${0.30 \pm 0.15}$ & ${0.43 \pm 0.05}$ & $\mathbf{0.60 \pm 0.04}$ \\
        & Sample Size  & ${0.31 \pm 0.12}$  & ${0.10 \pm 0.20}$ & ${0.20 \pm 0.07}$ & ${0.46 \pm 0.05}$ \\ 
        & PropScore Error  & ${0.21 \pm 0.09}$  & ${0.10 \pm 0.20}$ & ${0.30 \pm 0.07}$ & ${0.43 \pm 0.05}$ \\ 
        & CovDist  & ${0.03 \pm 0.06}$  & ${0.10 \pm 0.20}$ & ${0.160 \pm 0.04}$ & ${0.46 \pm 0.05}$ \\ 
        \midrule
        \multirow{5}{*}{Bayesian CF} & 
        $\mathrm{EIG}_{\theta_{\textit{c}|\cD_0 }}$             & $\mathbf{0.54 \pm 0.10}$  & $\mathbf{0.60 \pm 0.15}$  & $\mathbf{0.63 \pm 0.08}$ & $\mathbf{0.70 \pm 0.04}$ \\
        & $\mathrm{EIG}_{\theta|\cD_0}$       & $0.36 \pm 0.10$ & $0.30 \pm 0.14$  &  $0.50 \pm 0.07$ & $\mathbf{0.66 \pm 0.05}$ \\
        & PropScore Error  & $0.45 \pm 0.11$ & $\mathbf{0.60 \pm 0.14}$   & $\mathbf{0.63 \pm 0.08}$ & $\mathbf{0.70 \pm 0.04}$ \\ 
        & Sample Size  & ${0.16 \pm 0.09}$  & ${0.20 \pm 0.11}$ & ${0.26 \pm 0.06}$ & ${0.52 \pm 0.05}$ \\
        & CovDist  & ${0.07 \pm 0.09}$  & ${0.00 \pm 0.00}$ & ${0.26 \pm 0.06a}$ & ${0.46 \pm 0.05}$ \\ 
        \bottomrule
        \end{tabular}%
\end{table}


\begin{table}[h!]
\centering
\caption{Lalonde dataset ranking experiment results with 15 candidate datasets} \label{tab:Lalonde_results}
\begin{tabular}{llcccc}
        \toprule
              \textbf{Model} & \textbf{Objective} & $\boldsymbol{\rho} (\uparrow)$ & \textbf{p@1} $(\uparrow)$ & \textbf{p@3} $(\uparrow)$ & \textbf{p@5} $(\uparrow)$ \\ \midrule
        \multirow{5}{*}{Polynomial} &
         $\mathrm{EIG}_{\theta_{\textit{c}|\cD_0 }}$ & $\mathbf{0.47 \pm 0.05}$& $\mathbf{0.40 \pm 0.11}$& $\mathbf{0.60 \pm 0.05}$& $\mathbf{0.79 \pm 0.04} $\\
        & $\mathrm{EIG}_{\theta|\cD_0}$  & $\mathbf{0.43 \pm 0.05}$& $\mathbf{0.35 \pm 0.13}$& ${0.48 \pm 0.04}$& ${0.53 \pm 0.03}$\\
        & PropScore Error  & ${0.19 \pm 0.10}$& ${0.20 \pm 0.17}$& ${0.32 \pm 0.06}$& ${0.49 \pm 0.05}$\\
        & Sample Size  & ${0.24 \pm 0.10}$& ${0.25 \pm 0.08}$& ${0.38 \pm 0.08}$& ${0.58 \pm 0.06}$\\
        & CovDist  & ${0.20 \pm 0.04}$& ${0.25 \pm 0.07}$& ${0.38 \pm 0.06}$& ${0.48 \pm 0.07}$\\
        \midrule
        \multirow{5}{*}{Causal GP} & 
         $\mathrm{EIG}_{\Tilde{\tau}(X_0)|\cD_0 }$         & $\mathbf{0.42 \pm 0.07}$& $\mathbf{0.5 \pm 0.12}$& $\mathbf{0.55 \pm 0.05}$& $\mathbf{0.72 \pm 0.03} $\\
         & $\mathrm{EIG}_{\mathbf{f}|\cD_0}$        & $\mathbf{0.41 \pm 0.04}$& $\mathbf{0.4 \pm 0.1}$& ${0.43 \pm 0.04}$& ${0.58 \pm 0.07}$\\
        & PropScore Error  & ${0.19 \pm 0.05}$& ${0.21 \pm 0.15}$& ${0.32 \pm 0.06}$& ${0.43 \pm 0.07}$\\ 
        & Sample Size  & ${0.13 \pm 0.07}$& ${0.15 \pm 0.08}$& ${0.31 \pm 0.09}$& ${0.53 \pm 0.06}$\\ 
        & CovDist  & ${0.22 \pm 0.04}$& ${0.25 \pm 0.07}$& ${0.36 \pm 0.08}$& ${0.48 \pm 0.04}$\\ 
        \midrule
        \multirow{5}{*}{Bayesian CF} & $\mathrm{EIG}_{\theta_{\textit{c}|\cD_0 }}$   & $\mathbf{0.44 \pm 0.05}$ & $\mathbf{0.45 \pm 0.08}$& $\mathbf{0.55 \pm 0.05}$& $\mathbf{0.78 \pm 0.04} $ \\
        & $\mathrm{EIG}_{\theta|\cD_0}$       & $\mathbf{0.39 \pm 0.05}$ & $\mathbf{0.45 \pm 0.06}$&  ${0.43 \pm 0.04}$& ${0.52 \pm 0.03}$\\
        & PropScore Error  & ${0.22 \pm 0.06}$& ${0.2 \pm 0.07}$& ${0.32 \pm 0.06}$& ${0.47 \pm 0.07}$\\ 
        & Sample Size  & ${0.18 \pm 0.07}$& ${0.2 \pm 0.06}$& ${0.35 \pm 0.09}$& ${0.41 \pm 0.06}$\\
        & CovDist  & ${0.34 \pm 0.03}$& ${0.3 \pm 0.07}$& ${0.42 \pm 0.08}$& ${0.61 \pm 0.04}$\\ 
        \bottomrule
        \end{tabular}%
\end{table}

\newpage
\section{Related work: further details}\label{ap:further_related_work}

\paragraph{On Causal Federated Learning}
Federated learning is a distributed machine learning approach that enables multiple parties to collaboratively train a shared model while keeping their raw data decentralised and private. Various federated learning approaches have been proposed, including federated stochastic gradient descent \citep{shokri2015privacy}, federated averaging \citep{mcmahan2017communication}, and more recently, methods for joint learning of deep neural network models \citep{sattler2019robust, wang2020optimizing}. 
However, these algorithms do not inherently support causal inference, as the dissimilar distributions across different data sources may lead to biased causal effect estimation. To date, limited research has been conducted on the federated estimation of causal effects, highlighting the need for further exploration in this area. Due to the scope of our work, in the following paragraphs, we will focus on presenting Federated Learning methods for CATE estimation, where covariate distribution and treatment allocation are not assumed to be identical across datasets. \\
Several methods propose disentangling the loss function to facilitate federated learning. \citet{vo2022adaptive} propose CausalRFF, an adaptive kernel approach for causal inference that utilises Random Fourier Features to partition the loss function into multiple components, with each component corresponding to a specific data source. However, CausalRFF approach lacks strong privacy guarantees to prevent data recovery, and modeling complex non-linear relationships remains challenging \citep{almodovar2023federated}. \citet{liu2024disentangle} introduce a Bayesian method where parameters refer to a local disentangled loss and are updated cross-silo using server aggregation. Similarly, \citet{vo2023federated} divide the loss function into site-specific functions, and specify a variational posterior distribution for each local loss. 
Instead of tackling the loss function, \citet{almodovar2023federated}) introduce a method based on disentanglement of latent factors into instrumental, confounding, and risk factors, which are then used for treatment effect estimation. They apply federated averaging on a neural network-based generative causal inference model. 
Ultimately, FedCov \citep{tarumi2023personalized} is a parametric method for federated adjustment of covariate distributions between sites, where sample weights are derived from a propensity-like model. 
In all the aforementioned methods, the accuracy of causal estimation is reduced due to the constraints of federated learning. Conversely, our approach does not alter the causal estimation step, thereby maintaining optimal estimation accuracy. The framework we propose focuses on federated learning of the Expected Information Gain that would be obtained by merging with a dataset. 
While some Federated Causal Learning methods \citep{vo2022adaptive, vo2023federated} provide uncertainty bounds, which could potentially be used to decide which dataset to merge with by comparing the uncertainty in these bounds, the provided bounds apply to the federated estimate and not the causal estimate potentially obtained after merging. Ultimately, none of these methods provide strong privacy guarantees, such as differential privacy (DP), which would ensure that raw data cannot be recovered from the model parametrisation or summary statistics. Moreover, all these methods use the outcome values for training their federated model, and outcome values tend to be more sensitive in nature.

\paragraph{On Causal Differential Privacy}
Contrasting with previous approaches,  \citet{niu2022differentially} introduce a meta-algorithm that adds differential privacy (DP) guarantees to various popular CATE estimation frameworks, addressing the privacy concerns mentioned earlier. However, their method relies on multiple sample splitting, where separate subsets of the data are used for estimating the propensity score and the joint response model. This approach allows for parallel composition, a property of differential privacy. In contrast, our work prioritises data efficiency, and aims to utilise the entire dataset for CATE estimation without the need for sample splitting.

\paragraph{On Bayesian Experimental Design} 
Bayesian Active Learning by Disagreement (BALD) \citep{houlsby2011bayesian} is a method designed to strategically acquire training data by focusing on regions of high uncertainty.  BALD introduces an acquisition function rooted in information theory, which guides the data acquisition process. When reducing entropy towards all parameters in \autoref{subsec:background_and_cond_outcome}, we introduce a new setting for BALD where dataset are considered as data points. In the CausalBALD \citep{jesson2021causal} approach, the acquisition function is altered to specifically target areas where the distributions of different treatment groups overlap, thereby maximizing sample efficiency for learning personalised treatment effects. CausalBALD is also made for the acquisition of individual data points. However, contrarily to BALD, CausalBALD's acquisition function cannot provide a scalar measure if we compute it for a dataset (i.e.\ a matrix $\{\bx_i, t_i \}^{n_e}_{i=1}$) instead of data points (i.e.\ a vector $\bx_i, t_i$ for a given $i$). To apply CausalBALD in our setting, one would need to approximate the higher-order interaction terms between all combinations of data points within each dataset, thus making the computation intractable.


\end{document}